\documentclass[12pt]{article}

\usepackage{amsmath,amsthm,amssymb,amsopn,amsfonts}
\usepackage{graphics}
\usepackage{graphicx}
\usepackage{subfigure}

\usepackage{hyperref}

\newcommand{\twospace}{\renewcommand{\baselinestretch}{1.3}\normalsize}

\newcommand{\R}{\mathbb{R}}

\newcommand{\A}{{\mathcal A}}
\newcommand{\B}{{\mathcal B}}
\newcommand{\G}{{\mathcal G}}

\newtheorem{theorem}{Theorem}
\newtheorem{proposition}[theorem]{Proposition}
\newtheorem{lemma}[theorem]{Lemma}

\twospace
\begin{document}

\title{On the Incommensurability Phenomenon}
\author{Donniell E. Fishkind, Cencheng Shen, Youngser Park, Carey E. Priebe\\
Department of Applied Mathematics and Statistics, Johns Hopkins University}
\maketitle

\begin{abstract} Suppose that two large, multi-dimensional data sets are
each noisy measurements of the same underlying random process, and principle
components analysis is performed separately on the data sets
to reduce their dimensionality. In some circumstances it may
happen that the two lower-dimensional data sets have an
inordinately large Procrustean fitting-error between them. The purpose
of this manuscript is to quantify this ``incommensurability phenomenon."
In particular, under specified conditions, the square
Procrustean fitting-error of the two normalized
lower-dimensional data sets is (asymptotically)
a convex combination (via a correlation parameter) of the Hausdorff
distance between the projection subspaces and the maximum possible
value of the square Procrustean fitting-error for normalized data.
We show how this gives rise to the incommensurability phenomenon,
and we employ illustrative simulations and also use real data
to explore how the incommensurability phenomenon may have an
appreciable~impact.\\

\end{abstract}

\section{Overview and Outline}

The ever-increasing importance of modern big-data analytics brings with it
the imperative to understand fusion and and inference on multiple and
massive  disparate, distributed data sets. What processing can be profitably
done separately, for subsequent joint inference? In the case where each data
set consists of measurements on the same objects, and combining the full data
sets is prohibitively expensive, it seems reasonable to separately project
each large, high-dimensional collection to a low-dimensional space, and to
then combine the representations. Unfortunately, this model can lead to
undesirable incommensurability with significant deleterious effects on
fusion and inference. In this manuscript, we quantify an appearance of this
phenomenon.

In Section \ref{caut} we begin with an idealized Tale of Two Scientists
and its accompanying Theorem \ref{main}, in order to pave the way for
our main result, Theorem \ref{realmain}---stated and proved in
Section \ref{general}---wherein, under more general conditions,
an asymptotic relationship is given between the Procrustean
fitting-error of the two lower dimensional data sets and a
distance between the projections.

Then, in Section \ref{sims}, we perform simulation experiments and utilize real
data to illustrate and support our main result
Theorem \ref{realmain}, and we
use these simulations and real data
to explore the implications of Theorem \ref{realmain}; in
particular, when two correlated data sets are separately projected
to achieve dimension reduction, and  when there is an insufficient
spectral gap in the covariance structure at the projection-dimension
cutoff, then large projection distance may result between the projections
for the two data sets, and inordinately large Procrustean
fitting-error then follows.
This ``incommensurability phenomenon" was named in Priebe et al. \cite{cepetal}.

\subsection{Background, and an applied, take-away lesson}

Dimension reduction is often applied to data before subsequent inference. Principal components analysis (PCA) \cite{anderson}, \cite{JolliffePCABook} and multi-dimensional scaling \cite{BorgBook} are two traditional tools for data processing; the Big Data trend has motivated many recent advances in dimension reduction, such as nonlinear dimension reduction \cite{TenenbaumSilvaLangford2000}, \cite{SaulRoweis2000}, \cite{BelkinNiyogi2003}, sparse and robust PCA \cite{ZouHastie2006}, \cite{CandesLi2009}, \cite{WittenTibshiraniHastie2009}, etc., which all achieve good performance in their respective domains. For the purpose of this paper, we confine ourselves to principal components analysis, which remains a very popular and successful method for processing data.

Procrustean fitting-error is a simple---yet useful---statistic
for the comparison of two correlated sets of spatial or spatially-embedded data. To give just a few examples, see \cite{Sibson1978} and \cite{Sibson1979} where Procrustes fit
is used to assess the goodness-of-fit between two slightly
different spatial configurations projected to a lower dimensional
space by multi-dimensional scaling. Procrustes analysis
is similarly seen to be a valuable tool for manifold alignment in \cite{Luo99}, \cite{WangMahadevan2008}, \cite{WangEtal2012}, and also see \cite{GoldbergRitov2009}; indeed, Procrustes fit can be used to
compare manifold-based embedding algorithms.

Several factors may
contribute to the manifestation of the incommensurability
phenomenon when two correlated data sets are projected
to a lower dimension. One factor is the circumstance
where the two data seta are projected separately
when the dimension reduction
is performed. Another factor is the circumstance where the choice of
embedding dimension $d$ doesn't leave a sufficiently large gap between
the $d$'th and $d+1$'th  eigenvalues of the covariance matrix for the
data sets. These factors may combine to
allow substantial probability of having significant
distance between the separate projection subspaces,
which then causes an inordinately large Procrustean
fitting-error.

Of course, one remedy is simply not to do the projections
separately for the two data sets;
robust joint embedding schemes are available, such
as developed in \cite{WangEtal2012}, \cite{SharmaEtAl2012}, and \cite{cepetal}. Indeed, an easily used candidate is canonical correlation analysis (CCA)
\cite{Hot36}, \cite{Hardoon2004}, which can be extended
to situations where more than two data sets are being treated,
and CCA has good properties for subsequent inferential tasks \cite{Sun13a}, \cite{Sun13b}, \cite{ShenEtAl2014}. The incommensurability phenomenon can then
be avoided at the cost of the extra computation involved, although
this extra computation might be a significant burden when dealing
with a large volume of data in a distributed system.

Another possible remedy would be to choose the embedding dimension $d$
so as to maintain enough of a gap between the
$d$'th and $d+1$'th eigenvalues of the data sets' covariance matrix.
However, this remedy may actually throw out the baby with
the bath water; indeed, limiting the embedding dimension
to maintain a healthy eigengap may come at the expense of
additional signal that might be mined by the inclusion of
additional dimensions, if practical.

Besides the theoretical relationships proven
in this manuscript, a practical and
applied contribution of this manuscript is the
take-away lesson and awareness of
the potential danger in not doing joint embedding
(or similar tactics) in the course of dimension reduction with
correlated data sets. Indeed, in the sections that follow,
 we provide an illustrative
vision of what~could~go~wrong.

\section{A cautionary Tale of Two Scientists \label{caut}}

For this section only, we explore an idealized scenario for
the purpose of straightforward illustration; the general setting will be treated in Section \ref{general}. For this entire manuscript, a general background reference
for matrix analysis tools that we employ (e.g.~Procrustes fitting, singular value
decomposition, spectral and norm identities and inequalities such as Weyl's Theorem
for Hermitian matrices and Interlacing inequalities for Hermitian matrices)
is the classical text \cite{HornandJohnson} , background on the Grassmannian (e.g.~principal angles, Hausdorff distance) useful for our particular work
is easily accessible in \cite{ckl}, and background on principal components
analysis (PCA) can be found in \cite{anderson}.
A classical and broad textbook on the Grassmannian is \cite{chikuse}.

Suppose that two scientists each take daily measurements of $m$
features of a random process, where $m$ is a
large, positive integer. For each day $i=1,2,3,\ldots$, the first
scientist records her daily measurements as ${\bf X}^{(i)} \in \R^m$,
where ${\bf X}^{(i)}_j$ is her measurement of the $j$th feature,
and the second scientist records his daily measurements as
 ${\bf Y}^{(i)} \in \R^m$, where ${\bf Y}^{(i)}_j$ is his
 measurement of the $j$th feature, for $j=1,2,\ldots,m$.
 Although the two scientists want to record the same process, suppose
 that their measurements are made with some error, which
 we model in the  following manner.

There are three collections of
random variables $\{ {\bf Z}_j^{(i)} \}$, $\{ {\bf Z'}_j^{(i)} \}$,
and $\{ {\bf Z''}_j^{(i)} \}$, each over indices $i=1,2,3,\ldots$
and $j=1,2,\ldots,m$, such that these random variables are all collectively independent and identically distributed, and their common distribution has finite variance
$\alpha>0$. Suppose that the
random variables $\{ {\bf Z}_j^{(i)} \}$ are the signal feature values
associated with the process that the scientists would like to record,
and the random variables $\{ {\bf Z'}_j^{(i)} \}$ and $\{ {\bf Z''}_j^{(i)} \}$ are
confounding noise.
Let a real-valued ``measurement-accuracy" parameter $\gamma$ be fixed in the interval $[0,1]$. One scenario is that for each day $i=1,2,3,\ldots$ and feature $j=1,2,\ldots,m$, the first scientist's measurement ${\bf X}_j^{(i)}$ is a mixture
of ${\bf Z}_j^{(i)}$ and ${\bf Z'}_j^{(i)}$
with respective probabilities $\gamma$ and $1-\gamma$, and the second
scientist's measurement ${\bf Y}_j^{(i)}$ is a mixture of ${\bf Z}_j^{(i)}$ and ${\bf Z''}_j^{(i)}$ with respective probabilities
$\gamma$ and $1-\gamma$. A second scenario is that, instead, for each day
$i=1,2,3,\ldots$ and feature $j=1,2,\ldots,m$, \ \ ${\bf X}_j^{(i)}=
\gamma \cdot {\bf Z}_j^{(i)} + \sqrt{1-\gamma^2} \cdot{\bf Z'}_j^{(i)}$
and ${\bf Y}_j^{(i)} = \gamma \cdot {\bf Z}_j^{(i)} + \sqrt{1-\gamma^2} \cdot
{\bf Z''}_j^{(i)}$. The main result of Section~\ref{caut} is Theorem \ref{main},
which will hold in either of these two scenarios.
At one extreme, if $\gamma=1$, then the two scientists' measurements are
perfectly accurate and ${\bf X}^{(i)}={\bf Y}^{(i)}$ for all $i$.
At the other extreme, if $\gamma=0$, then the two scientists' measurements
are independent of each other.

For each positive integer $n$, denote by $X^{(n)}$ the matrix
$[ {\bf X}^{(1)} | {\bf X}^{(2)} | \cdots {\bf X}^{(n)}]
\in \R^{m \times n}$ consisting of the first scientist's measurements
over the first $n$ days, and denote by $Y^{(n)}$ the matrix
$[ {\bf Y}^{(1)} | {\bf Y}^{(2)} | \cdots {\bf Y}^{(n)}]
\in \R^{m \times n}$ consisting of the second scientist's measurements
over the~first~$n$~days.

Because the measurement vectors are in high-dimensional space~$\R^m$,
suppose the scientists project their respective measurement vectors
to the lower-dimensional space $\R^k$ for some smaller, positive integer~$k$.
This is done in the following manner.
Let $H_n=I_n-\frac{1}{n}J_n$ denote the centering matrix ($I_n$ and $J_n$
are, respectively, the $n \times n$ identity matrix and the matrix of all ones).
Suppose that the first scientist chooses a sequence
$\A^{(1)}, \A^{(2)},  \A^{(3)}, \ldots $ of random (or deterministic)
elements of the Grassmannian $\G_{k,m}$ (the space of all $k$-dimensional subspaces of $\R^m$), and suppose that the second scientist chooses a sequence
$\B^{(1)}, \B^{(2)},  \B^{(3)}, \ldots $ of random (or deterministic)
elements of the Grassmannian $\G_{k,m}$.
No assumptions are made on the distributions of these elements of the
Grassmannian or on their dependence/independence, but one example of
interest is where, for $n=1,2,3,\ldots$, $\A^{(n)},\B^{(n)} \in \G_{k,m}$
denote the respective $k$-dimensional subspaces to which principal components
analysis (PCA) projects $X^{(n)}H_n$ and $Y^{(n)}H_n$, respectively (and separately).
Let $P_{\A^{(n)}}$ denote the projection operator from $\R^m$ onto $\A^{(n)}$.
On each day $n$, the first scientist reports the scaled matrix
${\mathcal X}^{(n)}:=\frac{\sqrt{k}}{\| P_{\A^{(n)}}X^{(n)}H_n  \|_F} P_{\A^{(n)}} X^{(n)}H_n \in \R^{k \times n}$ to the Governing Board of Scientists, and the
second scientist reports the scaled matrix
${\mathcal Y}^{(n)}:=\frac{\sqrt{k}}{\| P_{\B^{(n)}}Y^{(n)}H_n  \|_F} P_{\B^{(n)}} Y^{(n)}H_n \in \R^{k \times n}$ to the Governing Board of Scientists, where
$\| \cdot \|_F$ denotes the Froebenius norm.
(Nota Bene: The specific choice of $\sqrt{k}$ in the
 scaling $\| {\mathcal X}^{(n)}\|_F = \| {\mathcal Y}^{(n)}\|_F  =\sqrt{k}$
is an innocuous notational convenience.)

Now, the Governing Board of Scientists wants to perform its own
check that the two scientists are indeed taking measurements reflecting
the same process. So the Governing Board of Scientists computes
the {\it Procrustean fitting-error}
$\epsilon( {\mathcal X}^{(n)},{\mathcal Y}^{(n)}):= \min _{Q
\in \R^{k \times k}:Q^TQ=I_k}\| Q {\mathcal X}^{(n)}- {\mathcal Y}^{(n)}\|_F$.
 It will later be seen (from Equation (\ref{scn})) that the square
 Procrustean fitting-error satisfies
$0 \leq \epsilon^2({\mathcal X}^{(n)},{\mathcal Y}^{(n)})
\leq 2k$; the Governing Board of Scientists
reasons that this square Procrustean fitting-error should be small
(negligible compared to $2k$) if indeed $\gamma$ is close
to $1$. Is this reasoning valid?

In the following, $d(\cdot , \cdot )$ denotes the Hausdorff distance (e.g.~see \cite{ckl}) on the
 Grassmannian $\G_{k,m}$; in particular, for any $\A,\B \in \G_{k,m}$,
 $d(\A,\B)=\sqrt{\sum_{i=1}^k(2 \sin \frac{\theta_i (\A,\B)}{2})^2 }$ where
 $\{ \theta_i(\A,\B) \}_{i=1}^k$ are the principal angles between $\A$ and $\B$.
Note that the square Hausdorff distance satisfies $0 \leq d^2(\A,\B) \leq 2k$.

\begin{theorem} \label{main}
Almost surely, $\epsilon^2({\mathcal X}^{(n)},{\mathcal Y}^{(n)})
- \Big[ (1-\gamma^2) \cdot 2k+
\gamma^2 \cdot d^2(\A^{(n)},\B^{(n)}) \Big] \rightarrow 0$ as $n \rightarrow \infty$.
\end{theorem}

The proof of Theorem \ref{main} is given later, in Section \ref{secmain}, as a special case of the more general Theorem \ref{realmain}.

Theorem \ref{main}
says that $\epsilon^2({\mathcal X}^{(n)},{\mathcal Y}^{(n)})$ asymptotically
becomes this convex combination (via $\gamma^2$) of $2k$ and
$d^2(\A^{(n)},\B^{(n)})$. In particular, if $\gamma$ is close to $0$,
which implies that the two scientists' measurements are
independent of each other, then indeed
$\epsilon^2({\mathcal X}^{(n)},{\mathcal Y}^{(n)})$ is close to its
maximum possible value $2k$, but if $\gamma$ is close to $1$, meaning that the
scientists' measurements are close to being the same as each other, we then
have $\epsilon^2({\mathcal X}^{(n)},{\mathcal Y}^{(n)})$ close to
$d^2(\A^{(n)},\B^{(n)})$. Is this square Hausdorff distance close to zero
when $\gamma$ is close to $1$?

In Section \ref{sims} we show that, in fact, if the (separate)
principal components analysis projections are used then this may not be the
case, and the square Hausdorff distance $d^2(\A^{(n)},\B^{(n)})$ might
even be close to its maximum possible value of
$2k$. By contrast, here if the two scientists both used the
simple-minded projection consisting of just taking the first $k$
coordinates of $\R^m$ and ignoring the rest of the coordinates,
then $d^2(\A^{(n)},\B^{(n)})=0$, in which case $\gamma$ close to $1$ would
indeed yield $\epsilon^2({\mathcal X}^{(n)},{\mathcal Y}^{(n)})$ close to $0$.

\section{The asymptotic relationship between Procrustean fitting-error and the projection distance \label{general} }

The main result of this section is the statement and proof of Theorem
\ref{realmain}. We begin with a description of a general setting and
a list of basic facts that will be used in the proof of Theorem~\ref{realmain}.

\subsection{Preliminaries and the general  setting  \label{setting}  }

From this point and on, we will consider a much more general setting than the
idealistic setting of Section~\ref{caut}. Suppose now that
${\bf X}^{(1)},{\bf X}^{(2)},{\bf X}^{(3)},\ldots
\in \R^m$ and ${\bf Y}^{(1)},{\bf Y}^{(2)},{\bf Y}^{(3)},\ldots
\in \R^m$ are random vectors (for convenience, let us denote
${\bf X} \equiv {\bf X}^{(1)},
{\bf Y} \equiv {\bf Y}^{(1)}$) such that the stacked
random vectors $ \bigl [\begin{smallmatrix} {\bf X}^{(1)} \\
{\bf Y}^{(1)} \end{smallmatrix} \bigr ],
\bigl [\begin{smallmatrix} {\bf X}^{(2)} \\
{\bf Y}^{(2)} \end{smallmatrix} \bigr ],
\bigl [\begin{smallmatrix} {\bf X}^{(3)} \\
{\bf Y}^{(3)} \end{smallmatrix} \bigr ],\ldots
 \in \R^{2m}$
are independent, identically distributed, with covariance matrix
\[ \textup{Cov}
\left [ \begin{array}{c} {\bf X} \\
{\bf Y} \end{array}
\right ] = \left [ \begin{array}{cc} \textup{Cov}({\bf X})
&   \textup{Cov}({\bf X},{\bf Y})  \\
\textup{Cov}({\bf Y},{\bf X}) & \textup{Cov}({\bf Y})
\end{array}
\right ]   \in \R^{2m \times 2m}.
\]
(We no longer require, in the manner of Section \ref{caut}, that ${\bf X}$ and
${\bf Y}$ have independent, nor identically distributed components, nor that they
arise as a mixture of other random variables in any particular way.) Assume that
Cov$({\bf X})$ and Cov$({\bf Y})$ are both nonzero matrices.

 Then define, for each positive integer $n$, random matrix
 $X^{(n)}:=[ {\bf X}^{(1)} | {\bf X}^{(2)} | \cdots {\bf X}^{(n)}]
\in \R^{m \times n}$ and $Y^{(n)}:=[ {\bf Y}^{(1)} | {\bf Y}^{(2)} |
\cdots {\bf Y}^{(n)}] \in \R^{m \times n}$. Let  $\A^{(n)}\in \G_{k,m}$
denote the $k$-dimensional subspace to which principal components
analysis (PCA) projects $X^{(n)}H_n$, and let $\B^{(n)} \in \G_{k,m}$
 denote the $k$-dimensional subspace to which principal components
analysis (PCA) projects  $Y^{(n)}H_n$ (these projections being done
separately). In the special case where
Cov$({\bf X})$ and Cov(${\bf Y})$ are scalar multiples of $I_m$, then
we will explicitly allow $\{ \A^{(n)} \}_{n=1}^\infty ,
\{ \B^{(n)} \}_{n=1}^\infty$ to be any sequences of elements in $\G_{k,m}$ whatsoever, deterministic~or~random.

It is useful to consider the projections $P_{\A^{(n)}}$ and $P_{\B^{(n)}}$ as
$m \times m$ symmetric, idempotent matrices (i.e.,~keep the ambient
coordinate system $\R^m$ for the projection's range) and, for each
$n=1,2,\ldots$, define
${\mathcal X}^{(n)}=\frac{\sqrt{k}}{\| P_{\A^{(n)}}X^{(n)}H_n  \|_F} P_{\A^{(n)}} X^{(n)}H_n  \in \R^{m \times n}$ and ${\mathcal Y}^{(n)}:=\frac{\sqrt{k}}{\| P_{\B^{(n)}}Y^{(n)}H_n  \|_F} P_{\B^{(n)}} Y^{(n)}H_n \in \R^{m \times n}$. (There is
no difference for our results and for the Procrustean fitting-error if,
as in Section~\ref{caut}, we instead treated $P_{\A^{(n)}}$ and
$P_{\B^{(n)}}$ as functions $\R^m \rightarrow \R^k$ with the coordinate
systems of $\A^{(n)}$ and $\B^{(n)}$, respectively, in which case
we have ${\mathcal X}^{(n)}$ and
${\mathcal Y}^{(n)}$ in $\R^{k \times n}$ instead of $\R^{m \times n}$.)

For any matrix $C \in \R^{m \times m}$ with only real-valued eigenvalues
(eg, symmetric matrices), let $\lambda_1(C) \geq \lambda_2(C) \geq \cdots
\geq \lambda_m(C)$ denote the eigenvalues of $C$.
For any matrix $C \in \R^{m \times n}$, let $\sigma_1(C) \geq \sigma_2(C) \geq
\cdots \geq \sigma_{\min \{m,n \}}(C)$ denote the singular values
of $C$. Recall that if $C$ is symmetric and positive semidefinite
(e.g., a covariance matrix) then $\lambda_i (C)=\sigma_i(C)$ for all $i=1,2,\ldots,m$,
and recall that, for any $C \in \R^{m \times n},D \in \R^{n \times m}$, the
nonzero eigenvalues of $CD$ are the same as the nonzero eigenvalues of $DC$.
For any $\A,\B \in \G_{k,m}$ (with projection matrices $P_\A,P_\B$) and all
$i=1,2,\ldots,m$, we thus have $\sigma^2_i (P_\A P_\B)=\lambda_i(P_\A P_\B
P_\B ^T P_\A ^T)=\lambda_i (P_\A P_\B P_\B P_\A) = \lambda_i (P_\A P_\A
P_\B P_\B) =\lambda_i (P_\A P_\B)$. In fact, $P_\A P_\B$ has at most
$k$ positive eigenvalues and at most $k$
positive singular values (the rest of the eigenvalues and the rest of the
singular values are  all zero) and, for all $i=1,2,\ldots,k$,
\begin{eqnarray} \label{angle}
   \sigma_i(P_\A P_\B) = \sqrt{\lambda_i(P_\A P_\B)} = \cos \theta_i (\A,\B),
\end{eqnarray}
where  $\{ \theta_i(\A,\B) \}_{i=1}^k$ are the principal angles
between $\A$ and $\B$.

For each $n=1,2,\ldots$, the {\it Hausdorff distance} 
$d({\A^{(n)}}{\B^{(n)}})$ is the nonnegative square root of
\begin{eqnarray} \label{thd}
d^2({\A^{(n)}},{\B^{(n)}}):= \sum_{i=1}^k2^2\sin^2
(\frac{\theta_i({\A^{(n)}},{\B^{(n)}})}{2}) =
\sum_{i=1}^k2(1-\cos \theta_i ( \A^{(n)},\B^{(n)}
 )).
\end{eqnarray}
It is clear that $0 \leq d^2({\A^{(n)}},{\B^{(n)}}) \leq 2k$.
We also define, for each $n=1,2,\ldots $, the quantity
\begin{eqnarray} \label{ethest}
\eth^2 ( \A^{(n)}, \B^{(n)} ) := \sum_{i=1}^k
2 \left (
1-\frac{1}{\frac{1}{k}\sum_{j=1}^k \sigma_j \left ( \textup{Cov}({\bf X},{\bf Y})
\right ) }
 \sigma_i
\Big( P_{\A^{(n)}} \textup{Cov}({\bf X},{\bf Y})P_{\B^{(n)}}
\Big)  \right ).
\end{eqnarray}
Later, in Proposition \ref{fbnd}, we will prove it always holds that
$0 \leq  \eth^2 ( \A^{(n)}, \B^{(n)} ) \leq 2k$.
Note that if Cov$({\bf X},{\bf Y})$ is a nonzero scalar multiple of $I_m$
then $\eth^2 ( \A^{(n)}, \B^{(n)} )$ is equal to $d^2({\A^{(n)}},{\B^{(n)}})$
and, in fact, if  Cov$({\bf X},{\bf Y})$ is the zero matrix then we will define
$\eth^2 ( \A^{(n)}, \B^{(n)} ) \equiv d^2({\A^{(n)}},{\B^{(n)}})$ (because, indeed,  $\frac{1}{\frac{1}{k}\sum_{j=1}^k \sigma_j \left ( \textup{Cov}({\bf X},{\bf Y})
\right ) }$ is not defined). For this reason, we like to view
$\eth^2 ( \A^{(n)}, \B^{(n)} )$ as a weighted form of the square Hausdorff distance.

For each $n=1,2,\ldots,$ the Procrustean fitting-error is defined to be
\begin{eqnarray} \label{proc}
\epsilon( {\mathcal X}^{(n)},{\mathcal Y}^{(n)}):= \min _{Q
\in \R^{m \times m}:Q^TQ=I_m}\| Q {\mathcal X}^{(n)}- {\mathcal Y}^{(n)}\|_F .
\end{eqnarray}
In fact, it holds that
\begin{eqnarray} \label{scn}
\epsilon^2 ({\mathcal X}^{(n)},{\mathcal Y}^{(n)})=\| {\mathcal X}^{(n)} \|_F^2
+ \| {\mathcal Y}^{(n)}   \|^2_F - 2 \sum_{i=1}^m\sigma_i({\mathcal Y}^{(n)}
{\mathcal X}^{(n)T})= 2k -  2 \sum_{i=1}^m\sigma_i({\mathcal Y}^{(n)}
{\mathcal X}^{(n)T}).
\end{eqnarray}

\subsection{The result  \label{secmain} }

Within the setting of Section \ref{setting}, we
now state and prove the main result of Section \ref{general}:
\begin{theorem} \label{realmain}
In the setting of Section \ref{setting}, it holds almost surely that
$$\epsilon^2({\mathcal X}^{(n)},{\mathcal Y}^{(n)})
- \Big[ (1- \rho) \cdot 2k+
\rho \cdot \eth^2 (\A^{(n)},\B^{(n)}) \Big] \rightarrow 0$$
as $n \rightarrow \infty$, where $\rho$ is defined as
$$\rho:= \frac{
\sum_{j=1}^k \sigma_j ( \textup{Cov}({\bf X},{\bf Y}))
}{ \sqrt{
\sum_{j=1}^k \sigma_j ( \textup{Cov}({\bf X}) )}
\sqrt{ \sum_{j=1}^k \sigma_j ( \textup{Cov}({\bf Y}))} }. $$
\end{theorem}

\noindent In Proposition \ref{sbnd} we prove that $0 \leq \rho \leq 1$.
To prove Theorem \ref{realmain} we first establish Lemmas~\ref{tech2}~and~\ref{next}:

\begin{lemma} \label{tech2} Almost surely,
$\textup{trace}\frac{1}{n-1}P_{\A^{(n)}}X^{(n)}
H_n H_n^TX^{(n)T}P^T_{\A^{(n)}} \rightarrow
\sum_{i=1}^k \sigma_i (\textup{Cov} ({\bf X}))$ as
$n \rightarrow \infty$.
\end{lemma}

\noindent {\bf Proof of Lemma \ref{tech2}:} For each $n=1,2,3,\ldots$,
let us consider a singular value decomposition
\begin{eqnarray*}
X^{(n)}H_n=U^{(n)} \Lambda^{(n)} V^{(n)T}
\end{eqnarray*}
where $U^{(n)} \in
\R^{m \times m}$ is orthogonal, $\Lambda^{(n)} \in \R^{m \times n}$ is
a ``diagonal" matrix, with nonnegative diagonals non-increasing along its main
diagonal, and $V^{(n)} \in \R^{n \times n}$ is orthogonal.
By the definition of PCA,
\begin{eqnarray*}
P_{\A^{(n)}}X^{(n)}H_n= U^{(n)} E \Lambda^{(n)} V^{(n)T},
\end{eqnarray*}
where $E \in \R^{m \times m}$ is the
diagonal matrix with its first $k$ diagonals $1$ and its remaining
diagonals~$0$.
Thus, the matrix
\begin{eqnarray*}
X^{(n)}H_nH_n^TX^{(n)^T}=U^{(n)}
\Lambda^{(n)} \Lambda^{(n)T}    U^{(n)T}
\end{eqnarray*} and the matrix
\begin{eqnarray*}
P_{\A^{(n)}}X^{(n)}H_nH_n^TX^{(n)^T}P_{\A^{(n)}}^T =U^{(n)} E
 \Lambda^{(n)}  \Lambda^{(n)T}  E \ U^{(n)T}
\end{eqnarray*}
share their $k$ largest eigenvalues, and the remaining $m-k$ eigenvalues
of the latter matrix are~$0$. By the Strong Law of Large Numbers,
almost surely $\frac{1}{n-1} X^{(n)}H_nH_n^T X^{(n)^T} \rightarrow \textup{Cov}({\bf X})$, hence we have $\textup{trace}\frac{1}{n-1}P_{\A^{(n)}}X^{(n)}
H_n H_n^TX^{(n)T}P^T_{\A^{(n)}} \rightarrow
\sum_{i=1}^k \lambda_i (\textup{Cov} ({\bf X}))=
\sum_{i=1}^k \sigma_i (\textup{Cov} ({\bf X})) $ as
$n \rightarrow \infty$.

Lastly, recall that we explicitly allow $\{ \A^{(n)}\}_{n=1}^\infty$
to be any elements of $\G_{k,m}$ in the special case
that Cov$({\bf X}) = \alpha \cdot I_m$ for some $\alpha >0$; indeed, in this
special case, note that by the boundedness of $\{ P_{\A^{(n)}} \}_{n=1}^\infty$
and the Strong Law of Large Numbers that,
as $n \rightarrow \infty$,
\begin{eqnarray*}
\textup{trace}\frac{1}{n-1}P_{\A^{(n)}}X^{(n)}
H_n H_n^TX^{(n)T}P^T_{\A^{(n)}} =  \ \ \ \ \ \ \ \ \ \ \ \ \ \ \ \ \
\ \ \ \ \ \ \ \ \ \ \ \ \ \ \ \ \ \ \ \ \ \ \ \ \ \
\ \ \ \ \ \ \ \ \ \ \ \ \ \ \ \ \ \ \ \ \ \ \ \ \ \
\\ \alpha \cdot
\textup{trace} P_{\A^{(n)}} + \textup{trace} P_{\A^{(n)}}
\left (  \frac{1}{n-1} X^{(n)}H_nH_n^TX^{(n)^T} -
\alpha \cdot I_m \right ) P^T_{\A^{(n)}} \rightarrow \alpha k
 =\sum_{i=1}^k \sigma_i (\textup{Cov} ({\bf X})) ,
\end{eqnarray*}
as desired. $\qed$

\begin{lemma} \label{next} For $i=1,2,\ldots,m$, almost surely
$$  \sigma_i^2({\mathcal Y}^{(n)}{\mathcal X}^{(n)T})-
\delta \cdot  \sigma^2_i (
 P_{\A^{(n)}} \textup{Cov}({\bf X},{\bf Y})P_{\B^{(n)}})  \rightarrow 0 $$
as $n \rightarrow \infty$, where $\delta:= \frac{1}{
\frac{1}{k} \sum_{j=1}^k \sigma_j ( \textup{Cov}({\bf X}) )
 \ \cdot \  \frac{1}{k} \sum_{j=1}^k \sigma_j ( \textup{Cov}({\bf Y}))}   $.
\end{lemma}

\noindent {\bf Proof of Lemma \ref{next}:}
For each $n=1,2,\ldots$, expand the expression
${\mathcal Y}^{(n)}{\mathcal X}^{(n)T}({\mathcal Y}^{(n)}{\mathcal X}^{(n)T})^T$
by the definitions to write it as
${\mathcal Y}^{(n)}{\mathcal X}^{(n)T}({\mathcal Y}^{(n)}{\mathcal X}^{(n)T})^T
= \phi^{(n)} \cdot \Phi^{(n)} $ where  $\phi^{(n)}$ and  $\Phi^{(n)}$ are defined by
\begin{eqnarray*}
\phi^{(n)}:= \frac{k^2}{\textup{trace}\frac{1}{n-1}P_{\B^{(n)}} Y^{(n)}H_nH_n^TY^{(n)T}P^T_{\B^{(n)}} \cdot
\textup{trace}\frac{1}{n-1}P_{\A^{(n)}} X^{(n)}
H_n H_n^TX^{(n)T}P^T_{\A^{(n)}}}  \in \R
\end{eqnarray*}
and
\begin{eqnarray*}
\Phi^{(n)}:=  P_{\B^{(n)}}\left ( \frac{1}{n-1}Y^{(n)}H_n  H^T_nX^{(n)T} \right ) P^T_{\A^{(n)}}  P_{\A^{(n)}} \left ( \frac{1}{n-1}X^{(n)}H_n H^T_n
Y^{(n)T}  \right ) P^T_{\B^{(n)}}  \in \R^{m \times m} .
\end{eqnarray*}
Define
$\Psi_{X,Y}^{(n)}:=\frac{1}{n-1}X^{(n)}H_nH_n^T Y^{(n)T}- \textup{Cov}(
{\bf X},{\bf Y})$; by the Strong Law of Large Numbers, almost
surely $\Psi_{X,Y}^{(n)} \rightarrow 0$ as $n \rightarrow \infty$.
Thus, by the subadditivity and submultiplicativity of
the norm, and by the boundedness of $\{ P_{\A^{(n)}} \}_{n=1}^\infty $
and $ \{ P_{\B^{(n)}} \}_{n=1}^\infty $, we have
almost surely that
\begin{eqnarray} \label{biggs}
\|  \Phi^{(n)} -  P_{\B^{(n)}} \textup{Cov}^T({\bf X},{\bf Y})  P^T_{\A^{(n)}}
 P_{\A^{(n)}} \textup{Cov}({\bf X},{\bf Y})  P_{\B^{(n)}}^T \|_F
 & = &
\|P_{\B^{(n)}} \Psi^{(n)T}_{X,Y}  P^T_{\A^{(n)}}
 P_{\A^{(n)}} \Psi^{(n)}_{X,Y}  P_{\B^{(n)}}^T \\
& + & \nonumber
P_{\B^{(n)}} \Psi^{(n)T}_{X,Y}  P^T_{\A^{(n)}}
 P_{\A^{(n)}} \textup{Cov}({\bf X},{\bf Y})  P_{\B^{(n)}}^T\\
& + & \nonumber
P_{\B^{(n)}} \textup{Cov}^T({\bf X},{\bf Y})  P^T_{\A^{(n)}}
 P_{\A^{(n)}} \Psi^{(n)}_{X,Y}  P_{\B^{(n)}}^T \|_F  \rightarrow 0,
\end{eqnarray}
as $n \rightarrow \infty$.
Now, by Lemma \ref{tech2} and the definition of $\phi^{(n)}$, almost surely
$\phi^{(n)}\rightarrow \delta$ as $n \rightarrow \infty$, hence by (\ref{biggs}) and
the boundedness of $\{ P_{\A^{(n)}} \}_{n=1}^\infty $ and $ \{ P_{\B^{(n)}} \}_{n=1}^\infty $, we have
almost surely that
\begin{eqnarray*}
\| \phi^{(n)}  \cdot
\Phi^{(n)} - \delta \cdot
 P_{\B^{(n)}} \textup{Cov}^T({\bf X},{\bf Y})  P^T_{\A^{(n)}}
 P_{\A^{(n)}} \textup{Cov}({\bf X},{\bf Y})  P_{\B^{(n)}}^T \|_F
 \ \ \ \ \ \ \ \ \ \ \ \ \ \ \ \ \ \ \ \ \ \ \ \ \ \ \ \ \ \ \ \ \ \
 \ \ \ \ \ \ \ \ \ \ \ \ \ \ \ \  \\
 \leq \| \phi^{(n)}  \Big(
\Phi^{(n)} -
 P_{\B^{(n)}} \textup{Cov}^T({\bf X},{\bf Y})  P^T_{\A^{(n)}}
 P_{\A^{(n)}} \textup{Cov}({\bf X},{\bf Y})  P_{\B^{(n)}}^T \Big)
 \|_F   \ \ \ \ \ \  \\
 + \| \Big( \phi^{(n)} - \delta \Big)
 \cdot  P_{\B^{(n)}} \textup{Cov}^T({\bf X},{\bf Y})  P^T_{\A^{(n)}}
 P_{\A^{(n)}} \textup{Cov}({\bf X},{\bf Y})  P_{\B^{(n)}}^T
 \|_F \rightarrow 0
 \end{eqnarray*}
as $n\rightarrow \infty$. Thus, by Weyl's Theorem for Hermitian matrices, for
each $i=1,2,\ldots,m$ we have almost surely that
\begin{eqnarray}
\Big|
\lambda_i \Big[ \phi^{(n)} \cdot \Phi^{(n)} \Big] - \lambda_i \Big[  \delta \cdot
\Big( P_{\A^{(n)}} \textup{Cov}({\bf X},{\bf Y})  P_{\B^{(n)}}^T \Big)^T
P_{\A^{(n)}} \textup{Cov}({\bf X},{\bf Y})  P_{\B^{(n)}}^T  \Big] \Big| \rightarrow 0
\end{eqnarray}
as $n\rightarrow \infty$, from which Lemma \ref{next} follows, after noting that
$P_{\B^{(n)}}$ is symmetric. $\qed$ \\

We are now able to prove the main result of this section, Theorem \ref{realmain}.\\

\noindent {\bf Proof of Theorem \ref{realmain}:} Let $\delta$ be as defined
in Lemma \ref{next}. Note that for any
nonnegative, bounded  real sequences $\{a^{(n)}\}_{n=1}^\infty $ and $\{b^{(n)}\}_{n=1}^\infty $, it holds\footnote{Indeed, because
$a^{(n)}$ and $b^{(n)}$ are bounded,
and since $|a^{(n)}-b^{(n)}|=
|\sqrt{a^{(n)}}-\sqrt{b^{(n)}}| \cdot |\sqrt{a^{(n)}}+\sqrt{b^{(n)}}|$,
we have that  $\sqrt{a^{(n)}}-\sqrt{b^{(n)}} \rightarrow 0$ implies
$a^{(n)}-b^{(n)} \rightarrow 0$. (Without the boundedness assumption
this implication may not hold.) Conversely, if  $\sqrt{a^{(n)}}-\sqrt{b^{(n)}}
\not \rightarrow 0$, then there exists $c>0$ such that
$| \sqrt{a^{(n_i)}}-\sqrt{b^{(n_i)}}  |\geq c$ for a subsequence, in
which case $|a^{(n)}-b^{(n)}|=
|\sqrt{a^{(n)}}-\sqrt{b^{(n)}}| \cdot |\sqrt{a^{(n)}}+\sqrt{b^{(n)}}| \geq c \cdot c$,
hence $a^{(n)}-b^{(n)} \not \rightarrow 0$.}
that $a^{(n)}-b^{(n)} \rightarrow
0$ if and only if $\sqrt{a^{(n)}}-\sqrt{b^{(n)}} \rightarrow 0$, as $n \rightarrow
\infty$. Thus, by Lemma \ref{next}, and noting that the rank of
$P_{\A^{(n)}} \textup{Cov}({\bf X},{\bf Y})  P_{\B^{(n)}}$ is at most $k$,
we have almost surely that, as $n \rightarrow \infty$,
\begin{eqnarray} \label{point}
\sum_{i=1}^m\sigma_i ( {\mathcal Y}^{(n)}{\mathcal X}^{(n)T}) - \sqrt{\delta}
\cdot \sum_{i=1}^k  \sigma_i
(P_{\A^{(n)}} \textup{Cov}({\bf X},{\bf Y})P_{\B^{(n)}}) \rightarrow 0.
\end{eqnarray}
But the expression in (\ref{point}) can be simplified, by (\ref{scn})
and (\ref{ethest}), as
\begin{eqnarray*}
& &  2k- 2 \sum_{i=1}^m \sigma_i ( {\mathcal Y}^{(n)}{\mathcal X}^{(n)T})
- \Big[ 2k-2 \sqrt{\delta} \sum_{i=1}^k
 \sigma_i
(P_{\A^{(n)}} \textup{Cov}({\bf X},{\bf Y})P_{\B^{(n)}})
\Big]   \\
& = &  \epsilon^2 ( {\mathcal X}^{(n)},{\mathcal Y}^{(n)}  )-
\Big[ 2k- 2 \rho \sum_{i=1}^k
\left (
\frac{1}{\frac{1}{k}\sum_{j=1}^k \sigma_j \left ( \textup{Cov}({\bf X},{\bf Y})
\right ) }
 \sigma_i
\Big( P_{\A^{(n)}} \textup{Cov}({\bf X},{\bf Y})P_{\B^{(n)}}
\Big)  \right )
\Big]   \\
& = & \epsilon^2 ( {\mathcal X}^{(n)},{\mathcal Y}^{(n)}  ) -
\Big[ (1-\rho) \cdot 2k + \rho \cdot
\sum_{i=1}^k
2 \left (
1-\frac{1}{\frac{1}{k}\sum_{j=1}^k \sigma_j \left ( \textup{Cov}({\bf X},{\bf Y})
\right ) }
 \sigma_i
\Big( P_{\A^{(n)}} \textup{Cov}({\bf X},{\bf Y})P_{\B^{(n)}}
\Big)  \right )
\Big]\\
& = & \epsilon^2 ( {\mathcal X}^{(n)},{\mathcal Y}^{(n)}  ) -
\Big[ (1-\rho) \cdot 2k + \rho \cdot \eth^2 ( \A^{(n)}, \B^{(n)} )
\Big],
\end{eqnarray*}
which establishes Theorem \ref{realmain}. $\qed$\\

There is a special case of Theorem \ref{realmain} that
deserves attention:

\begin{theorem} \label{special} In the setting
of Section \ref{setting}, if $\textup{Cov}({\bf X})=\textup{Cov}({\bf Y})$ and
$\textup{Cov}({\bf X},{\bf Y})=\beta I_m$ for a real number $\beta$, then it holds
almost surely that
$$\epsilon^2({\mathcal X}^{(n)},{\mathcal Y}^{(n)})
- \Big[ (1- \frac{| \beta |}{ \alpha' } ) \cdot 2k+
\frac{| \beta |}{ \alpha' }
 \cdot d^2 (\A^{(n)},\B^{(n)}) \Big] \rightarrow 0$$
as $n \rightarrow \infty$, where $\alpha':=\frac{1}{k} \sum_{j=1}^k \sigma_j ( \textup{Cov}({\bf X}) )$.
\end{theorem}

\noindent
Theorem \ref{special} is an immediate consequence of Theorem \ref{realmain},
since we previously pointed out that when $\textup{Cov}({\bf X},{\bf Y})$ is
a scalar multiple of the identity then
$\eth^2 (\A^{(n)},\B^{(n)})   = d^2 (\A^{(n)},\B^{(n)})$. $\qed$\\

Finally, Theorem \ref{main} from Section \ref{caut}
is an immediate consequence of Theorem \ref{special}, after noting that
the setting of Section \ref{caut} is a special case of
the setting of Section \ref{setting}, with (recall the definitions of
$\alpha$ and $\gamma$ from Section \ref{caut})
\[ \textup{Cov}
\left [ \begin{array}{c} {\bf X} \\
{\bf Y} \end{array}
\right ] = \left [ \begin{array}{rr} \alpha \cdot I_m \ \ \
&   \gamma^2 \cdot \alpha \cdot I_m  \\
\gamma^2 \cdot \alpha \cdot I_m  \ \ \  & \alpha \cdot I_m
\end{array}
\right ]   \in \R^{2m \times 2m}.
\]
So $| \beta |$ and $\alpha'$ of Theorem \ref{special} are, respectively,
$\gamma^2 \cdot \alpha $ and $\alpha$, thus in Theorem \ref{special} we have
$\frac{|\beta |}{\alpha '} = \frac{\gamma^2 \cdot \alpha }{\alpha} = \gamma^2$.
This proves Theorem \ref{main}. $\qed$

\subsection{Bounds for $\eth^2$ and $\rho$}

\begin{proposition} \label{fbnd}
For  $\eth^2 ( \A^{(n)}, \B^{(n)} )$ as defined in (\ref{ethest}),
it  holds that $0 \leq \eth^2 ( \A^{(n)}, \B^{(n)} ) \leq 2k$.
\end{proposition}

\noindent
{\bf Proof of Proposition \ref{fbnd}:} The upper bound is trivial. To prove
the lower bound, first we re-express (\ref{ethest}) as
\begin{eqnarray} \label{boon}
\eth^2 ( \A^{(n)}, \B^{(n)} ) = \frac{2}{\frac{1}{k}\sum_{j=1}^k \sigma_j \left ( \textup{Cov}({\bf X},{\bf Y})
\right ) }    \sum_{i=1}^k
 \left ( \sigma_i (\textup{Cov}({\bf X},{\bf Y})) -  \sigma_i
\Big( P_{\A^{(n)}} \textup{Cov}({\bf X},{\bf Y})P_{\B^{(n)}}
\Big)  \right ),
\end{eqnarray}
and we show that each summand in the summation of  (\ref{boon}) will be
nonnegative. Indeed, for any
$S \in \R^{m \times m}$ and $i=1,2,\ldots, n$, we have that
$\sigma_i(S \cdot P_{\A^{(n)}}) \leq
\sigma_i(S)$ and $\sigma_i( P_{\A^{(n)}} S) \leq
\sigma_i(S)$; this is seen as follows.
Say $P_{\A^{(n)}}=QEQ^T$ is such that $Q \in \R^{m \times m}$ is orthogonal and
$E$ is diagonal with $1$'s and $0$'s on its diagonal. Then
$\sigma_i^2 (S \cdot P_{\A^{(n)}} )=\lambda_i ( P^T_{\A^{(n)}}
S^T S P_{\A^{(n)}}) = \lambda_i(QEQ^T S^TS Q E Q^T)=
\lambda_i(EQ^T S^TS Q E ) \leq \lambda_i(Q^T S^TS Q )
= \lambda_i (S^TS) = \sigma_i^2 (S)$, the inequality holding by
the Interlacing Theorem for Hermitian matrices. By a similar argument
 $\sigma_i( P_{\A^{(n)}} S) \leq \sigma_i(S)$, and applying these in
 succession yields that
 $  \sigma_i (P_{\A^{(n)}} \textup{Cov}({\bf X},{\bf Y})P_{\B^{(n)}})
 \leq \sigma_i (\textup{Cov}({\bf X},{\bf Y}))$.
 $\qed$

\begin{proposition} For $\rho$, as defined in Theorem \ref{realmain}, it
holds that $0 \leq \rho \leq 1$. \label{sbnd}
\end{proposition}

\noindent
{\bf Proof of Proposition \ref{sbnd}:} Let
$\textup{Cov}({\bf X},{\bf Y})=U \Lambda V^T$ be a singular
value decomposition; i.e. $U,V \in \R^{m \times m}$ are orthogonal and
$\Lambda \in \R^{m \times m}$ is diagonal, with nonincreasing
nonnegative diagonal entries. Define
$M \in \R^{2m \times 2m}$ by
\[ M:=
\left [ \begin{array}{cc} U^T
&  0_m  \\ 0_m & V^T
\end{array}
\right ]
\left [ \begin{array}{cc} \textup{Cov}({\bf X})
&   \textup{Cov}({\bf X},{\bf Y})  \\
\textup{Cov}^T({\bf X},{\bf Y}) & \textup{Cov}({\bf Y})
\end{array}
\right ]
\left [ \begin{array}{cc} U
&  0_m  \\ 0_m & V
\end{array}
\right ] =
\left [ \begin{array}{cc} U^T  \textup{Cov}({\bf X}) U
&  \Lambda  \\
\Lambda & V^T \textup{Cov}({\bf Y}) V
\end{array}
\right ]
\]
where $0_m \in \R^{m \times m}$ is the matrix of zeros. A covariance
matrix is positive semidefinite, thus $M$ is positive semidefinite,
as well as all of its principal submatrices.
For each $j=1,2,\ldots,k$, the two-by-two submatrix consisting of the
$j$th and $j+m$th rows and columns of $M$ has nonnegative
diagonals and a nonnegative determinant, thus
$( U^T  \textup{Cov}({\bf X}) U )_{jj}
 ( V^T  \textup{Cov}({\bf Y}) V )_{jj} \geq (\Lambda_{jj})^2$, i.e.
\begin{eqnarray}  \label{lat}
\sigma_j( \textup{Cov}({\bf X},{\bf Y}))  \leq
\sqrt{\Big( U^T  \textup{Cov}({\bf X}) U \Big)_{jj}}
\cdot \sqrt{\Big( V^T  \textup{Cov}({\bf Y}) V \Big)_{jj}}  .
\end{eqnarray}
Now, summing (\ref{lat}) over $j=1,2,\ldots,k$ and applying the
Cauchy-Schwartz inequality to the resulting right-hand side, we
obtain
\begin{eqnarray} \label{lata}
\sum_{j=1}^k \sigma_j( \textup{Cov}({\bf X},{\bf Y}))  \leq
\sqrt{ \sum_{j=1}^k \Big( U^T  \textup{Cov}({\bf X}) U \Big)_{jj}}
\cdot \sqrt{\sum_{j=1}^k \Big( V^T  \textup{Cov}({\bf Y}) V \Big)_{jj}}.
\end{eqnarray}
For any Hermitian matrix, the vector of its diagonals always majorizes the vector
of its eigenvalues, thus
\begin{eqnarray} \label{latb}
\sum_{j=1}^k \Big( U^T  \textup{Cov}({\bf X}) U \Big)_{jj} \leq
\sum_{j=1}^k \lambda_j (U^T  \textup{Cov}({\bf X}) U) =
\sum_{j=1}^k \sigma_j (\textup{Cov}({\bf X})),
\end{eqnarray}
and Proposition \ref{sbnd} follows from (\ref{lata}), (\ref{latb}), and
(\ref{latb}) applied to $\textup{Cov}({\bf Y})$ and $V$.
$\qed$

\subsection{An isometry-corrective property of $\eth^2$ \label{corrsect}}

Suppose that $W \in \R^{m \times m}$ is an orthogonal matrix such that
\[ \textup{Cov}
\left [ \begin{array}{r} {\bf X} \\
W{\bf Y} \end{array}
\right ] = \left [ \begin{array}{cc} \textup{Cov}({\bf X})
&  \beta \cdot I_m  \\
\beta \cdot I_m & \textup{Cov}({\bf X})
\end{array}
\right ]   \in \R^{2m \times 2m},
\]
where $\beta \in \R$ is nonzero; this might arise in situations
similar to the cautionary Tale of Two Scientists in Section \ref{caut}---wherein two scientists are taking measurements of the same random
process---except that the second scientist permutes the order of the
features (i.e., $W$ is a permutation matrix).
Define $W \B^{(n)}:=\{Wx: x \in \B^{(n)}\}$.
In this situation,
the quantity $d^2(\A^{(n)},W\B^{(n)})$ may be more
interesting~than~the~quantity $d^2(\A^{(n)},\B^{(n)})$, since
$\A^{(n)}$ might be viewed as being more comparable to $W\B^{(n)}$
then to $\B^{(n)}$.
Indeed, if the eigenvalues of Cov$({\bf X})$ are distinct
and $n$ is large and  $W$ is not $I_m$, then $d^2(\A^{(n)},W\B^{(n)})$
would be small, in contrast to $d^2(\A^{(n)},\B^{(n)})$.
\begin{proposition} \label{corrective}
In the case of the previous paragraph, we have
$\eth^2(\A^{(n)},\B^{(n)})=d^2(\A^{(n)},W\B^{(n)}) $.
\end{proposition}

\noindent
Proposition \ref{corrective} will be illustrated in Section \ref{illu}.\\

\noindent {\bf Proof of Proposition \ref{corrective}:}
Here we have
\[ \textup{Cov}
\left [ \begin{array}{r} {\bf X} \\
{\bf Y} \end{array}
\right ] =
\left [ \begin{array}{cc} I_m & 0_m \\ 0_m & W^T  \end{array} \right ]
\left [ \begin{array}{cc} \textup{Cov}({\bf X})
&  \beta \cdot I_m  \\
\beta \cdot I_m & \textup{Cov}({\bf X})
\end{array}
\right ]
\left [ \begin{array}{cc} I_m & 0_m \\ 0_m & W  \end{array} \right ]
=
\left [ \begin{array}{cc} \textup{Cov}({\bf X})
&  \beta \cdot W  \\
\beta \cdot W^T & \textup{Cov}({\bf Y})
\end{array}
\right ]  ,
\]
thus for all $i=1,2,\ldots,m$
\begin{eqnarray}
& & \sigma_i(\textup{Cov} ({\bf X},{\bf Y}))=\sigma_i (\beta \cdot W)
=|\beta | \label{ft}  \\  \label{fs} & \mbox{ and } &
\sigma_i(P_{\A^{(n)}} \textup{Cov} ({\bf X},{\bf Y})
P_{\B^{(n)}}) =  |\beta| \cdot \sigma_i(P_{\A^{(n)}} W P_{\B^{(n)}})
=   |\beta | \cdot \sigma_i(P_{\A^{(n)}} W P_{\B^{(n)}}W^T)
\end{eqnarray}
Because $P_{W\B^{(n)}}=W P_{\B^{(n)}}W^T$, and by
(\ref{thd}), (\ref{ethest}), (\ref{ft}), and (\ref{fs})
 it follows that \\
$\eth^2(\A^{(n)},\B^{(n)})=\sum_{i=1}^k2(1-
\sigma_i(P_{\A^{(n)}} W P_{\B^{(n)}}W^T))=d^2(\A^{(n)},W \B^{(n)})$.
$\qed$

\section{Simulations and Real Data \label{sims}}

In this section, simulations and real data illustrate and support
the theorems which we stated and proved in the previous sections,
and we then use these simulations and real data to illustrate how the
``incommensurability phenomenon" can arise as a consequence. What is meant
by this phenomenon is the occurrence an inordinately large Procrustean
fitting-error between projected data that was originally highly-correlated.
(This phenomenon was named in Priebe et al \cite{cepetal}.)

\subsection{A first illustration  \label{firill}}

Our first illustration of Theorem \ref{realmain} and Theorem \ref{special}
is with ${\bf X}$ and ${\bf Y}$ distributed
multivariate normal (with mean vector consisting
of all zeros) such that Cov$({\bf X})=$Cov$({\bf Y})=I_6$ and
Cov$({\bf X},{\bf Y})= \beta \cdot I_6$ for assorted values of $\beta$.
Note that $\rho$ as defined in Theorem \ref{realmain}
is  $\beta$ here, note that  $\alpha'$ and $\beta$ as defined in
Theorem \ref{special} are, respectively, $1$ and $\beta$ here,
and note that  here
 $\eth^2(\A^{(n)},\B^{(n)})=d^2(\A^{(n)},\B^{(n)})$
 because Cov$({\bf X},{\bf Y})$ is a scalar
 multiple of the identity.
Also, this example may be seen as an illustration of
Theorem~\ref{main}---in the Tale of Two Scientists---with
$\gamma^2$ there being $\beta$ here.

The dimension of the space containing
${\bf X}$ and ${\bf Y}$ is $m=6$, and we will project to spaces of dimension $k=2$.

\begin{figure}
 \centering
 \mbox{\subfigure{\includegraphics[height=5in,width=3.5in]{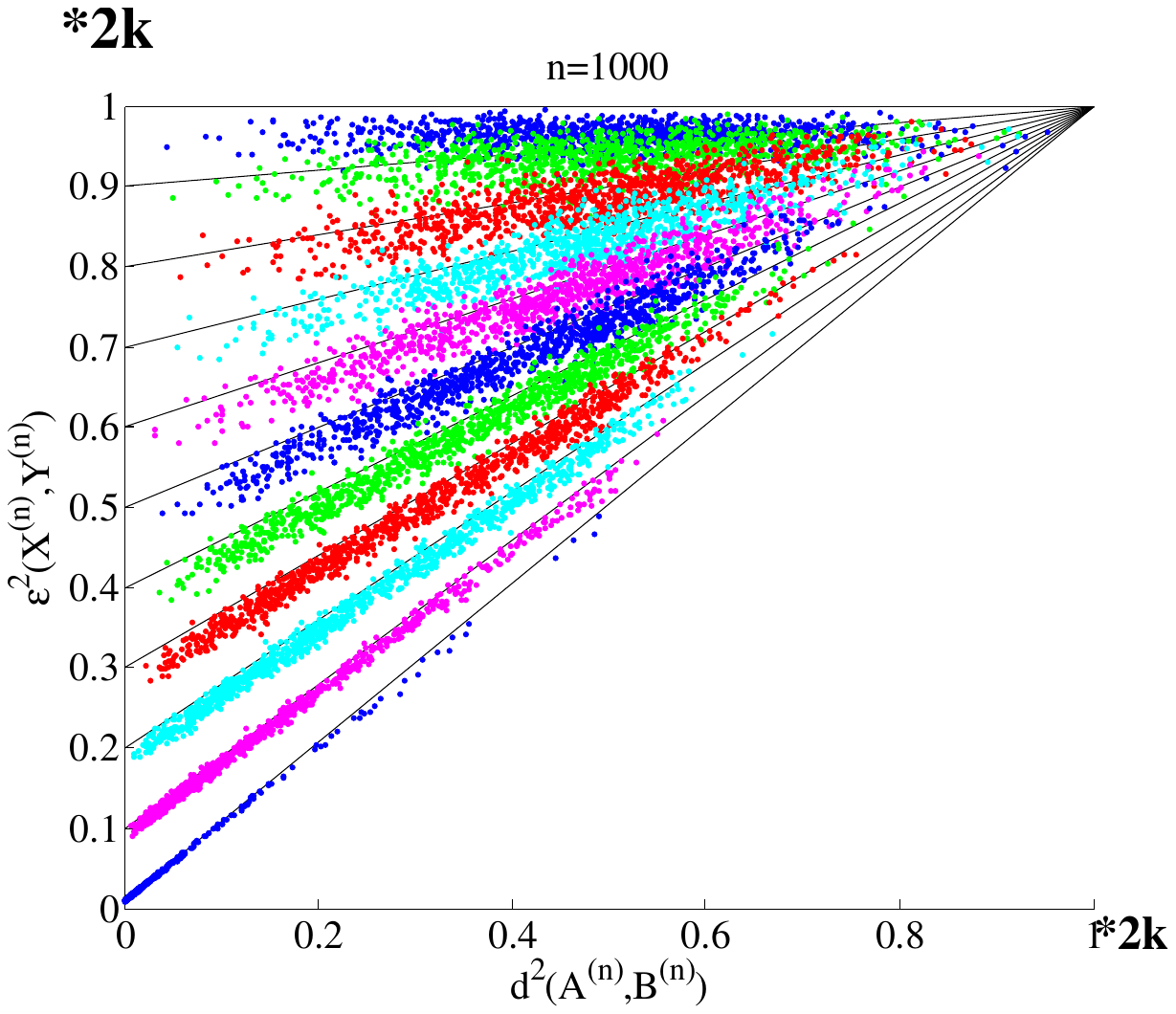}
 \quad
 \subfigure{\includegraphics[height=5in,width=3.5in]{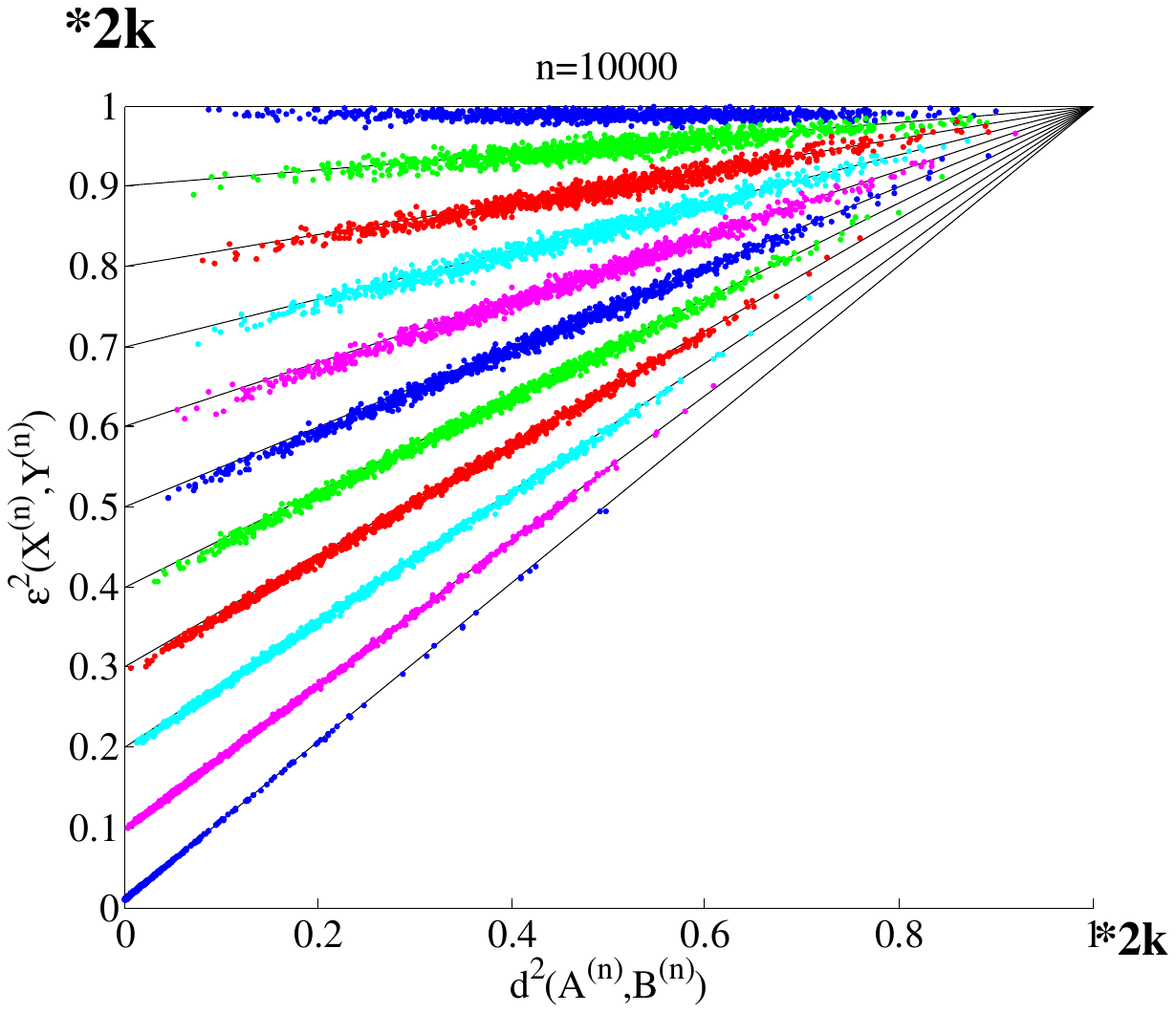} }}}
 \caption{Plots of $\epsilon^2({\mathcal X}^{(n)},{\mathcal Y}^{(n)})$
vs $d^2 (\A^{(n)},\B^{(n)})$ when Cov$({\bf X})=$Cov$({\bf Y})=I_6$,
Cov$({\bf X},{\bf Y})= \beta \cdot I_6$.
For each of $\beta =0$ (blue), $.1$ (green), $.2$ (red), $.3$ (cyan),~$.4$~(magenta), $.5$ (blue), $.6$ (green), $.7$ (red), $.8$ (cyan), $.9$ (magenta), $.99$ (blue), for each of $n=1000$ (left) and $n=10000$ (right), there were $1000$ Monte Carlo replicates
using $k=2$. Note that the axis-values are to be
multiplied by $2k$, which is $4$ here, since the ranges of
$\epsilon^2({\mathcal X}^{(n)},{\mathcal Y}^{(n)})$ and $d^2 (\A^{(n)},\B^{(n)})$
are the interval $[0,2k]$.}
 \label{fig1}
 \end{figure}

For each of $\beta=0,.1,.2,.3,.4,.5,.6,.7,.8,.9,.99$, and for each of
$n=1000$ and $n=10000$ we obtained $1000$ realizations of
${\mathcal X}^{(n)}$ and ${\mathcal Y}^{(n)}$ and used
PCA to obtain $\A^{(n)}$ and $\B^{(n)}$. In Figure~\ref{fig1},
we plotted the values of  $\epsilon^2({\mathcal X}^{(n)},{\mathcal Y}^{(n)})$
against the respective values of $d^2 (\A^{(n)},\B^{(n)})$,
in colors blue, green, red, cyan, magenta, blue, green, red, cyan, magenta, blue
for the respective values of  $\beta=0,.1,.2,.3,.4,.5,.6,.7,.8,.9,.99$.
For reference, we also included---in Figure \ref{fig1}---lines
with y-intercept $(1- \beta) \cdot 2k$
and slope $\beta$, for each of the above-specified values of
$\beta$; basically,
Theorem~\ref{main}, Theorem~\ref{realmain}, and Theorem~\ref{special}
state that the scatter plots will
adhere to these respective lines in the limit as $n$ goes to $\infty$.
Indeed, notice in Figure \ref{fig1} that the scatter plots  adhere very closely
to their respective lines, and such adherence  substantially improves as
$n=1000$ is raised to $n=10000$, which supports/illustrates the claims of
Theorem \ref{main}, Theorem \ref{realmain}, and Theorem~\ref{special}.

The above was done using
PCA to generate $\A^{(n)}$ and $\B^{(n)}$. What if we
instead took $\A^{(n)}$ and $\B^{(n)}$ to (each) be the span
of the first two standard-basis vectors in $\R^6$? We will
call this the ``trivial" choice of $\A^{(n)}$ and $\B^{(n)}$.
Of course, the value of $d^2 (\A^{(n)},\B^{(n)})$ would always be
identically zero, and note that
Theorem \ref{main}, Theorem \ref{realmain}, and Theorem~\ref{special}
still apply with this choice of $\A^{(n)}$ and $\B^{(n)}$
because Cov$({\bf X})$ and Cov$({\bf Y})$ are scalar multiples
of the identity. Thus, the scatter plots from these above experiments
when they are performed instead for the trivial choice
of $\A^{(n)}$ and $\B^{(n)}$ would land in the far left of
Figure \ref{fig1} (along the y-axis at $d^2 (\A^{(n)},\B^{(n)})=0$),
clustered about their respective lines.
Indeed, we then performed the above experiments for the trivial choice of
$\A^{(n)}$ and $\B^{(n)}$;
the sample  mean and sample standard deviation
of $\epsilon^2({\mathcal X}^{(n)},{\mathcal Y}^{(n)})$
for the $1000$ Monte Carlo replicates when $n=10000$ were as follows:
\[
\begin{array}{c||c|c}
& \mbox{mean, st.dev.~of $\epsilon^2({\mathcal X}^{(n)},{\mathcal Y}^{(n)})$ with PCA}
&  \mbox{mean, st.dev.~of $\epsilon^2({\mathcal X}^{(n)},{\mathcal Y}^{(n)})$ with trivial $\A^{(n)}$ and $\B^{(n)}$} \\ \hline
\beta=0  & 3.9546, \ \ 0.0170  & 3.9534, \ \ 0.0178  \\
\beta=.1 & 3.7903, \ \ 0.0618  & 3.6003, \ \ 0.0277  \\
\beta=.2 & 3.5774, \ \ 0.1179  & 3.1994, \ \ 0.0276 \\
\beta=.3 & 3.3413, \ \ 0.1796  & 2.7990, \ \ 0.0254 \\
\beta=.4 & 3.0942, \ \ 0.2429  & 2.4006, \ \ 0.0230 \\
\beta=.5 & 2.7918, \ \ 0.3043  & 1.9999, \ \ 0.0210   \\
\beta=.6 & 2.4581, \ \ 0.3658  & 1.5996, \ \ 0.0177 \\
\beta=.7 & 2.0331, \ \ 0.4283  & 1.2007, \ \ 0.0140 \\
\beta=.8 & 1.5368, \ \ 0.4567  & 0.8003, \ \ 0.0103 \\
\beta=.9 & 0.9232, \ \ 0.4607  & 0.4001, \ \ 0.0054 \\
\beta=.99 & 0.1352, \ \ 0.2057 & 0.0400, \ \ 0.0006
\end{array}
\]
Indeed, besides the notable exception when $\beta=0$ (where there is
no correlation anyway between ${\bf X}$ and ${\bf Y}$),
the values of $\epsilon^2({\mathcal X}^{(n)},{\mathcal Y}^{(n)})$
were substantially larger when PCA was used
to generate $\A^{(n)}$ and $\B^{(n)}$
than for the trivial choice of $\A^{(n)}$ and $\B^{(n)}$.
This is the incommensurability phenomenon, a situation
where use of PCA has the consequence of inordinately large
Procrustean fitting-error.

Let us call the values $\epsilon^2({\mathcal X}^{(n)},{\mathcal Y}^{(n)})
- \Big[ (1-\beta) \cdot 2k+
\beta \cdot d^2(\A^{(n)},\B^{(n)}) \Big]$ {\it residuals}. It is noteworthy
that in the above experiments the sample standard deviation of the residuals
when PCA was used to generate $\A^{(n)}$ and $\B^{(n)}$ is very close
to the sample standard deviation of $\epsilon^2({\mathcal X}^{(n)},{\mathcal Y}^{(n)})$
for the trivial choice of $\A^{(n)}$ and $\B^{(n)}$. Specifically, we computed:
 \[
\begin{array}{c||c|c}
& \mbox{st.dev.~of residuals with PCA}
&  \mbox{st.dev.~of $\epsilon^2({\mathcal X}^{(n)},{\mathcal Y}^{(n)})$ with trivial $\A^{(n)}$ and $\B^{(n)}$} \\ \hline
\beta=0  &  0.0170 & 0.0178  \\
\beta=.1 &  0.0267 & 0.0277  \\
\beta=.2 &  0.0262 & 0.0276 \\
\beta=.3 & 0.0252 & 0.0254 \\
\beta=.4 &  0.0235 & 0.0230 \\
\beta=.5 &  0.0214 & 0.0210   \\
\beta=.6 &  0.0192 & 0.0177 \\
\beta=.7 &  0.0158 & 0.0140 \\
\beta=.8 & 0.0118 & 0.0103 \\
\beta=.9 &  0.0073 & 0.0054 \\
\beta=.99 &  0.0025 & 0.0006
\end{array}
\]
So, it seems empirically here that
the variation in $\epsilon^2({\mathcal X}^{(n)},{\mathcal Y}^{(n)})$
not explained by $d^2(\A^{(n)},\B^{(n)})$ when PCA generates
$\A^{(n)}$ and $\B^{(n)}$ is approximately the same as the
variation in $\epsilon^2({\mathcal X}^{(n)},{\mathcal Y}^{(n)})$ for
the trivial choice of $\A^{(n)}$ and $\B^{(n)}$ (in which
$d^2(\A^{(n)},\B^{(n)})=0$ identically) and, as such,
$d^2(\A^{(n)},\B^{(n)})$ explains all of the rest
of the variation here in
$\epsilon^2({\mathcal X}^{(n)},{\mathcal Y}^{(n)})$ when PCA is used.

\subsection{A second illustration  \label{secillu}}

Our next illustration of Theorem \ref{realmain} and
Theorem \ref{special} is with ${\bf X}$ and ${\bf Y}$ multivariate normal
(with mean vector of all zeros) such that Cov$({\bf X})=$Cov$({\bf Y})=$
the diagonal matrix in $\R^{20 \times 20}$ with $.7$ on all diagonals except
for the first diagonal, which has the value $1$, and such that
Cov$({\bf X},{\bf Y})=.6 * I_{20}$. So we are using $m=20$ here.
As above, $\eth^2(\A^{(n)},\B^{(n)})=d^2(\A^{(n)},\B^{(n)})$
because Cov$({\bf X},{\bf Y})$ is a scalar multiple of the identity.

We will use three different projection dimensions, each of $k=1,2,10$.
When $k=1$ the formula in Theorem \ref{realmain} yields
$\rho=\frac{.6}{1}=.6$, when $k=2$ the formula yields
$\rho=\frac{.6+.6}{1+.7}\approx .7059$, and when
$k=10$ the formula yields $\rho=\frac{.6+.6+.6 \cdots }{1+.7+.7 \cdots}\approx
.8219$.

\begin{figure}
 \centering
 \mbox{\subfigure{\includegraphics[height=5in,width=3.5in]{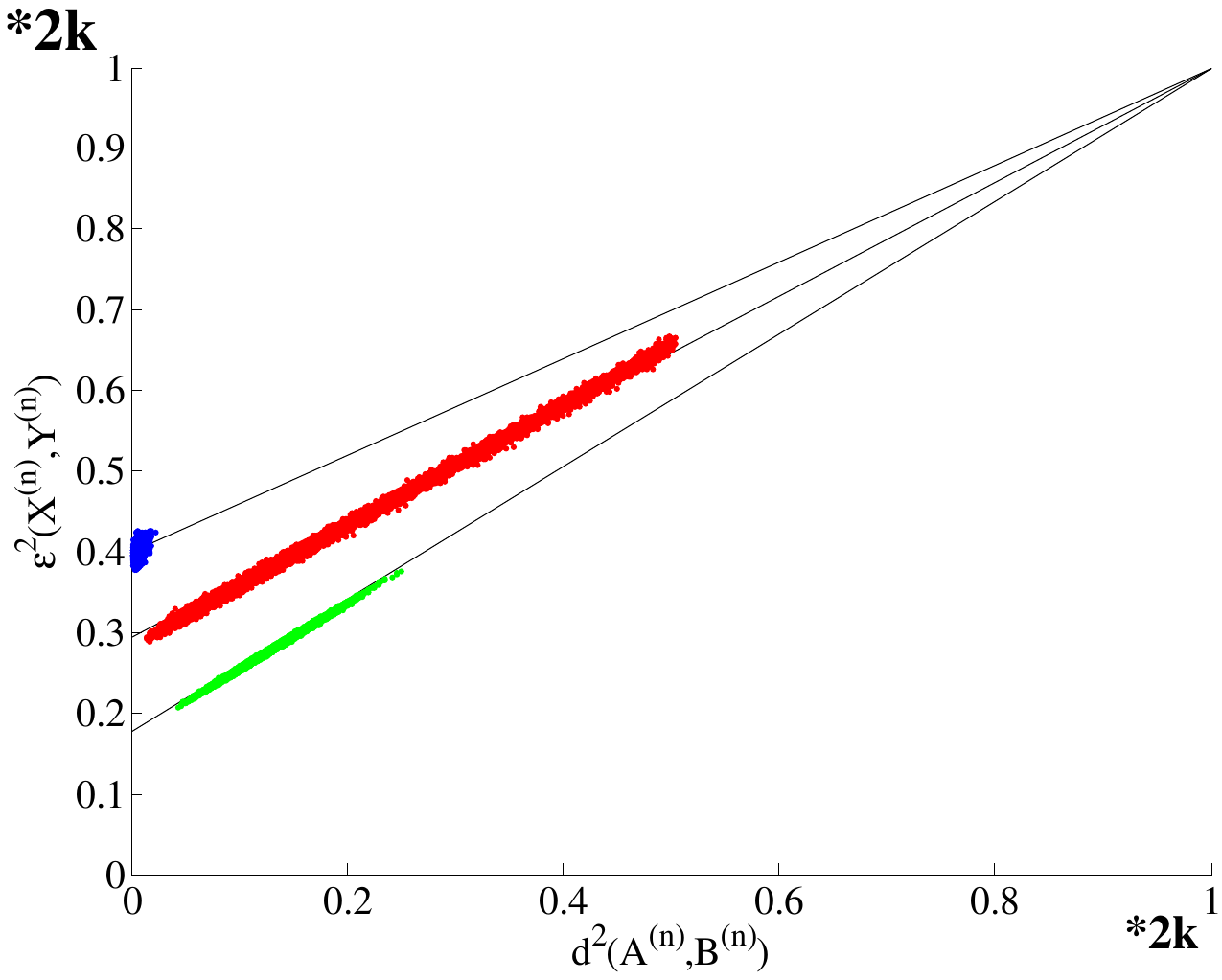}
 \quad
 \subfigure{\includegraphics[height=5in,width=3.5in]{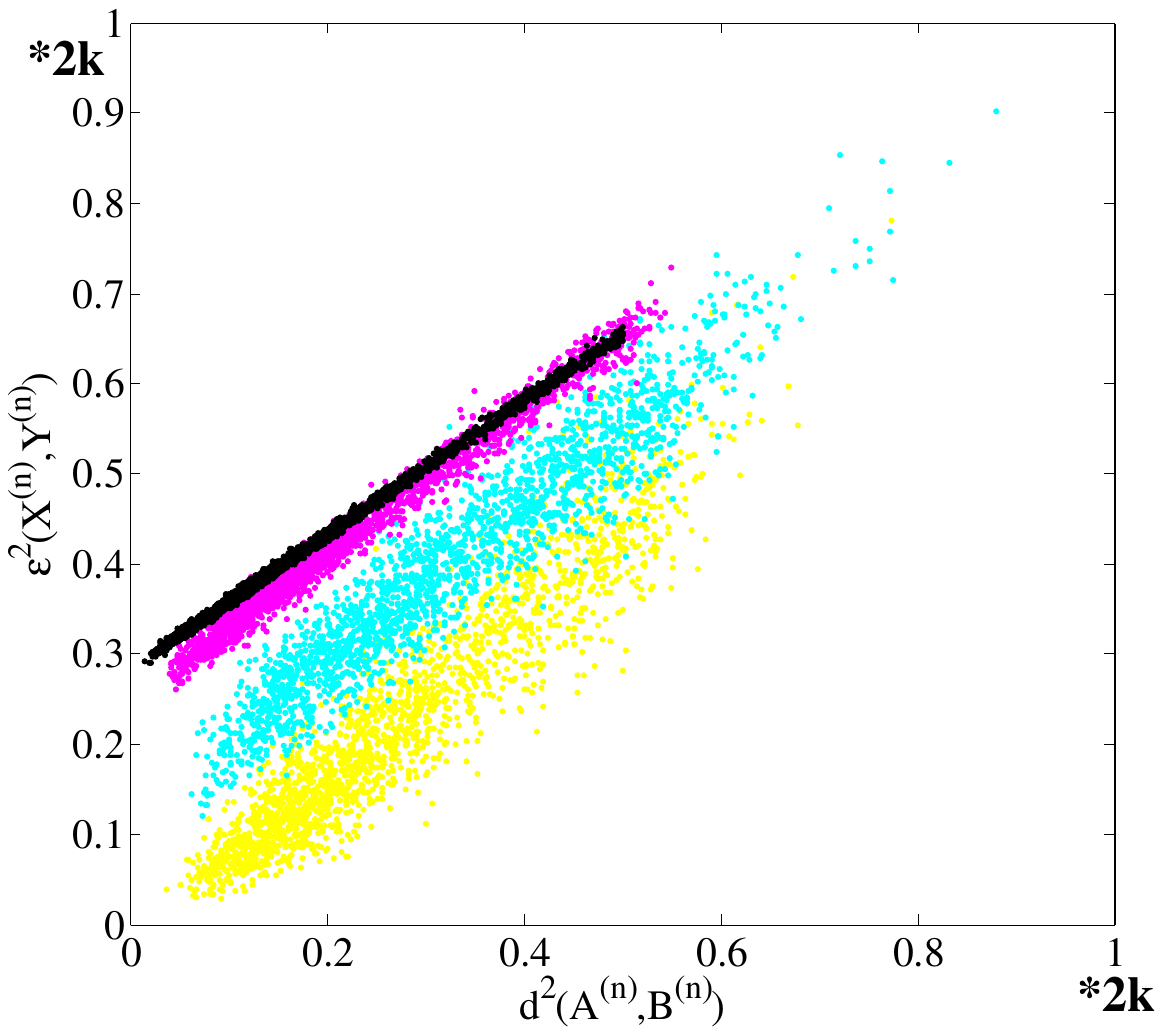} }}}
 \caption{Plots of $\epsilon^2({\mathcal X}^{(n)},{\mathcal Y}^{(n)})$
vs $d^2 (\A^{(n)},\B^{(n)})$ for
Cov$({\bf X})=\textup{Cov}({\bf Y})=diag(1,.7,.7,\ldots,.7)\in \R^{20 \times 20}$,
Cov$({\bf X},{\bf Y})= .6* I_{20}$.
The figure on the left shows $10000$ Monte Carlo
replications when $n=10000$, for each of $k=1$ (blue), $k=2$ (red), and $k=10$ (green). The figure on the right shows $2000$ Monte Carlo replications when
$k=2$, for each of $n=10^1$ (yellow), $n=10^2$ (cyan), $n=10^3$ (magenta), and
$n=10^4$ (black).
Note that the axis-values are to be
multiplied by $2k$ for the respective values of $k$,
 since the ranges of $\epsilon^2({\mathcal X}^{(n)},{\mathcal Y}^{(n)})$ and $d^2 (\A^{(n)},\B^{(n)})$ are  the interval~$[0,2k]$.}
 \label{fig2}
 \end{figure}

Using PCA to generate $\A^{(n)}$ and $\B^{(n)}$,
we obtained $10000$ realizations
of ${\mathcal X}^{(n)}$ and ${\mathcal Y}^{(n)}$ when $n=10000$, for each
projection dimension $k=1$, $k=2$, and $k=10$; the values of
$\epsilon^2({\mathcal X}^{(n)},{\mathcal Y}^{(n)})$ are plotted
against the respective values of $d^2 (\A^{(n)},\B^{(n)})$ in the left figure
of Figure \ref{fig2}, with $k=1$ in blue, $k=2$ in red, and $k=10$ in green.
As before, lines are drawn on the figure to indicate the limiting relationship between
 $\epsilon^2({\mathcal X}^{(n)},{\mathcal Y}^{(n)})$ and
 $d^2 (\A^{(n)},\B^{(n)})$ that is predicted by
 Theorem~\ref{realmain} and Theorem~\ref{special}; indeed, the scatter
 plots adhere very closely to these respective lines.
 In the right hand side of Figure \ref{fig2} is $2000$ Monte Carlo
 simulations when $k=2$ for each of $n=10^1$ (yellow), $n=10^2$ (cyan),
 $n=10^3$ (magenta) and $n=10^4$ (black). As $n$ is getting larger,
 these are seen to get increasingly closer to the corresponding
 limiting relationship between
 $\epsilon^2({\mathcal X}^{(n)},{\mathcal Y}^{(n)})$ and
 $d^2 (\A^{(n)},\B^{(n)})$. All of this supports the claims of
 Theorem \ref{realmain} and Theorem \ref{special}.

In the experiments for the left figure in Figure \ref{fig2},
the sample mean and sample standard deviation of
$\frac{\epsilon^2({\mathcal X}^{(n)},{\mathcal Y}^{(n)})}{2k}$
were as follows:
\[
\begin{array}{c||c|c}
 & \mbox{sample mean of } \frac{\epsilon^2({\mathcal X}^{(n)},{\mathcal Y}^{(n)})}{2k} & \mbox{sample standard deviation of } \frac{\epsilon^2({\mathcal X}^{(n)},{\mathcal Y}^{(n)})}{2k} \\ \hline
 k=1 & .4017   &  .0069   \\
 k=2 & .4323   &  .0950   \\
 k=10 & .2797   & .0244    \\
\end{array}
\]
(We normalize $\epsilon^2({\mathcal X}^{(n)},{\mathcal Y}^{(n)})$
with division by $2k$ since the range of
$\epsilon^2({\mathcal X}^{(n)},{\mathcal Y}^{(n)})$ is $[0,2k]$.
As $k$ increases, the correlation $\rho$ increases, so it would seem
at first thought that the normalized Procrustean fitting-error
$\frac{\epsilon^2({\mathcal X}^{(n)},{\mathcal Y}^{(n)})}{2k}$ should
decrease. Indeed, the leftmost green points in (the left figure of)
Figure \ref{fig2} are below the leftmost red points, which are below
the leftmost blue points. However, overall, the normalized Procrustean
fitting-error is seen in the table above to be
much higher in the case of $k=2$ than the
case of $k=1$. This is explained by noting a substantial gap between
the first eigenvalue of Cov$({\bf X})$ and the second eigenvalue of Cov$({\bf X})$
($1$ vs $.7$) whereas there is no gap between the second eigenvalue of
of Cov$({\bf X})$ and the third eigenvalue of Cov$({\bf X})$ (both are $.7$). Thus when
$k=1$ the PCA projection has little variance whereas when $k=2$ the
PCA projection has much variance, often causing much larger
Hausdorff distance between $\A^{(n)}$ and $\B^{(n)}$, which results in
larger Procrustean fitting-error by Theorem \ref{realmain}. As such,
the case of $k=2$  is an example of the incommensurability
phenomenon of inordinately large Procrustean fitting-error.
But then observe that when $k=10$ we find that the normalized
Procrustean fitting-error is competitive with the $k=1$ case;
even though the tenth and  eleventh eigenvalues of Cov$({\bf X})$ are the same,
nonetheless the correlation $\rho$ has increased, and the variance of
the PCA projection has decreased enough to improve the normalized
Procrustean fitting-error to be competitive with the case of $k=1$.

Not only may the incommensurability phenomenon occur
when there is no spectral gap in the covariance structure
at the projection dimension, but the incommensurability phenomenon
may occur when this spectral gap is positive but small.
Indeed, repeating the experiments performed for the left figure in Figure \ref{fig2},
and just changing the second diagonal of Cov$({\bf X})=$Cov$({\bf Y})$ from
$.7$ to $\lambda$ for each of $\lambda=.71, .72, .73, .74, .75$
but otherwise the experiments are the same, we got a very
similar-looking scatter plot as the left figure in Figure
\ref{fig2}, and the sample mean and sample standard
deviation of $\frac{\epsilon^2({\mathcal X}^{(n)},{\mathcal Y}^{(n)})}{2k}$
were as follows:
\[
\begin{array}{c||c|c|c}
  & k=1 & k=2 & k=10 \\ \hline \hline
\mbox{sample mean of $\frac{\epsilon^2({\mathcal X}^{(n)},{\mathcal Y}^{(n)})}{2k}$when $\lambda=.71$} & .4017 & .4342 & .2807 \\
\mbox{sample mean of $\frac{\epsilon^2({\mathcal X}^{(n)},{\mathcal Y}^{(n)})}{2k}$when $\lambda=.72$} & .4018 & .4329 & .2809 \\
\mbox{sample mean of $\frac{\epsilon^2({\mathcal X}^{(n)},{\mathcal Y}^{(n)})}{2k}$when $\lambda=.73$} & .4018 & .4207 & .2810 \\
\mbox{sample mean of $\frac{\epsilon^2({\mathcal X}^{(n)},{\mathcal Y}^{(n)})}{2k}$when $\lambda=.74$} & .4018 & .3997 & .2815\\
\mbox{sample mean of $\frac{\epsilon^2({\mathcal X}^{(n)},{\mathcal Y}^{(n)})}{2k}$when $\lambda=.75$} & .4019 & .3754 & .2815 \\ \hline
\mbox{sample stdev of $\frac{\epsilon^2({\mathcal X}^{(n)},{\mathcal Y}^{(n)})}{2k}$when $\lambda=.71$} & .0068 & .0960 & .0244 \\
\mbox{sample stdev of $\frac{\epsilon^2({\mathcal X}^{(n)},{\mathcal Y}^{(n)})}{2k}$when $\lambda=.72$} & .0070 & .0953 & .0243 \\
\mbox{sample stdev of $\frac{\epsilon^2({\mathcal X}^{(n)},{\mathcal Y}^{(n)})}{2k}$when $\lambda=.73$} & .0069 & .0922 & .0245 \\
\mbox{sample stdev of $\frac{\epsilon^2({\mathcal X}^{(n)},{\mathcal Y}^{(n)})}{2k}$when $\lambda=.74$} & .0070& .0832 & .0244 \\
\mbox{sample stdev of $\frac{\epsilon^2({\mathcal X}^{(n)},{\mathcal Y}^{(n)})}{2k}$when $\lambda=.75$} & .0070 & .0662 & .0242
\end{array}
\]
In the case of $k=2$, the spectral gap in the covariance structure at the
projection dimension is $\lambda - .7$, and note that as this gap grows to
$.75 - .7 =.05$  there is a lessening of the incommensurability phenomenon,
but the phenomenon is still very much present. Indeed (from the table above),
when $\lambda=.75$, the sample mean of $\frac{\epsilon^2({\mathcal X}^{(n)},
{\mathcal Y}^{(n)})}{2k}$
when $k=2$ (see table above) is below the sample mean when $k=1$, but it is
only lower by less than a half of the sample standard deviation of
$\frac{\epsilon^2({\mathcal X}^{(n)},{\mathcal Y}^{(n)})}{2k}$ when
$k=2$ and, in fact, notice that the sample standard deviation of
$\frac{\epsilon^2({\mathcal X}^{(n)},{\mathcal Y}^{(n)})}{2k}$ when
$k=2$ is more than $9$ times the sample standard deviation when $k=1$.
Thus there is a significant probability of an inordinately high
Procrustean fitting error in the case of $k=2$ with $\lambda=.75$.

\subsection{A modification of the second illustration to illustrate the
isometry-corrective property of $\eth^2$ \label{illu}}

Our next illustration of Theorem \ref{realmain} is with ${\bf X}$ and
${\bf Y}$ distributed multivariate normal, with joint covariance matrix given by:

\[
\textup{Cov}
\left [ \begin{array}{r} {\bf X} \\
{\bf Y} \end{array}
\right ]
= \left [
\begin{array}{cccccccccccc}
1 & 0 & 0 & \hdots & 0 & 0   & 0 & 0 & \hdots & 0 & 0 & .6 \\
0 & .7 & 0 & \hdots & 0 & 0 & 0 & 0 & \hdots & 0 & .6 & 0\\
0 & 0 & .7 & \hdots & 0 & 0 & 0 & 0 & \hdots & .6 & 0 & 0\\
\vdots & \vdots & \vdots & \ddots & \vdots & \vdots & \vdots & \vdots &
 & \vdots & \vdots & \vdots \\
0 & 0 & 0 & \hdots & .7 & 0 & 0 & .6 & \hdots & 0 & 0 & 0\\
0 & 0 & 0 & \hdots & 0 & .7 & .6 & 0 & \hdots & 0 & 0 & 0 \\
0 & 0 & 0 & \hdots & 0 & .6 & .7 & 0 & \hdots & 0 & 0 & 0 \\
0 & 0 & 0 & \hdots & .6 & 0 & 0 & .7 & \hdots & 0 & 0 & 0 \\
\vdots & \vdots & \vdots &  & \vdots & \vdots & \vdots & \vdots &
\ddots & \vdots & \vdots & \vdots \\
0 & 0 & .6 & \hdots & 0 & 0 & 0 & 0 & \hdots & .7 & 0 & 0 \\
0 & .6 & 0 & \hdots & 0 & 0 & 0 & 0 & \hdots & 0 & .7 & 0 \\
.6 & 0 & 0 & \hdots & 0 & 0 & 0 & 0 & \hdots & 0 & 0 & 1
\end{array}  \right ]    \in \R^{40 \times 40}.
\]

Of course, this is exactly the illustration in the
beginning of Section \ref{secillu},
with the only exception that the coordinates of $Y$ have been permuted
into reverse order. Performing the very same experiments
from the beginning of
Section \ref{secillu}, the scatter plots of
$\epsilon^2({\mathcal X}^{(n)},{\mathcal Y}^{(n)})$
vs $d^2 (\A^{(n)},\B^{(n)})$ will {\bf not} look like
the scatter plots in Figure \ref{fig2}.
However, since the permutation transformation is an isometry, we then have
by Proposition \ref{corrective} in Section \ref{corrsect}, that
the scatter plots of $\epsilon^2({\mathcal X}^{(n)},{\mathcal Y}^{(n)})$
vs $\eth^2 (\A^{(n)},\B^{(n)})$ {\bf will indeed}
look like the scatter plots in Figure \ref{fig2}. The use of
$\eth^2$ automatically accounts for isometrical transformations of
${\bf X}$ and/or ${\bf Y}$ from a common frame, in the manner of this
example.

It should also be mentioned that, for the illustration of this section
(with the covariance matrix above),
if $\A^{(n)}$ and $\B^{(n)}$ were not generated with PCA, but instead
$\A^{(n)}$ and $\B^{(n)}$ were selected to be (the same as each other by setting
them to be) the span of any number of standard-basis vectors
in $\R^{40}$ then the Procrustes fitting-error
would be disasterously large. The fact that such a naive choice of
$\A^{(n)}$ and $\B^{(n)}$ was successful in the illustration
in Section \ref{firill} was just a byproduct of the good fortune
that ${\bf X}$ and ${\bf Y}$ did not have permuted coordinates or
any other isometrical transformation applied to them.

\subsection{The incommensurability phenomenon in real data   \label{realdata}}

We next illustrate the incommensurability phenomenon
using real data from the 2014 {\it Science} article of
Vogelstein et.~al.~\cite{Science}, titled
``Discovery of brainwide neural-behavioral maps via
multiscale unsupervised structure learning."
See  O'Leary and Marder \cite{ScienceAbout} for a
big-picture overview and discussion of the contributions
of this article.
This data from Vogelstein et.~al.~will be observed by
Two Scientists who will record highly correlated observations.
We will show the incommensurability phenomenon creeping into
the Two Scientist's efforts. (The data related to this section 
is available online at 
\url{http://www.cis.jhu.edu/~parky/Incomm/}.)

\begin{figure}
 \centering
\includegraphics[height=10in,width=8in]{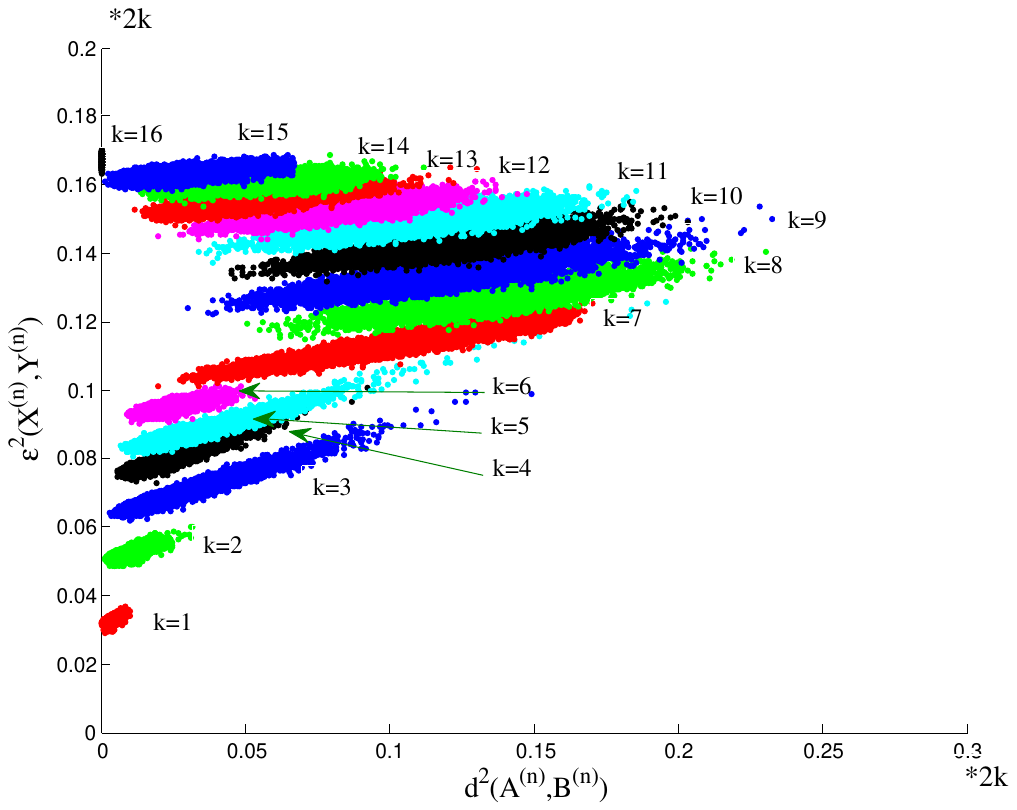} \vspace{-2.7in}
\caption{An illustration of the incommensurability phenomenon.
Plots of $\epsilon^2({\mathcal X}^{(242)},{\mathcal Y}^{(242)})$
vs $d^2 ({\mathcal A}^{(242)},{\mathcal B}^{(242)})$ for
10000 Monte-Carlo replicates using the Drosophila larvae data,
for each embedding dimension $k=1,2,3,\ldots, 16$. Notice the change
from $k=6$ to $k=7$.
(Also note that the axis-values are to be
multiplied by $2k$ for the respective values of $k$,
 since the ranges of $\epsilon^2({\mathcal X}^{(242)},{\mathcal Y}^{(242)})$ and $d^2 ({\mathcal A}^{(242)},{\mathcal B}^{(242)})$ are  the interval~$[0,2k]$).}
 \label{fignice}
 \end{figure}

In the Vogelstein et.~al.~paper \cite{Science}, the authors consider a
collection of optogenetically manipulated Drosophila larvae,
with the goal of generating a behavioral reference atlas.
The animals considered are partitioned into {\em lines}, with each line
defined by the neuron classes which are being optogenetically manipulated.
Each line includes multiple replicates -- {\em dishes} -- in each of which
numerous animals are found. Videos of animal behavior are processed into
a multivariate behavioral time series for each animal, these time series give
rise to an animal dissimilarity matrix, and multidimensional scaling applied
to this dissimilarity matrix yields a representation of the collection of
animals in high-dimensional Euclidean space. For our purposes, we will focus
our attention on the sixteen most significant dimensions; in this manner,
every animal corresponds to a vector in $\R^{16}$.

In our experiment here, for each of $i=1,2,3,\ldots,242$, the first scientist
will record ${\bf X}^{(i)}\in \R^{16}$ and the second scientist will record
${\bf Y}^{(i)}\in \R^{16}$, as follows. There were a total of
$n=242$ dishes in the Vogelstein et.~al.~data set
corresponding to the control line {\em pBDPU-ChR2}.
For each $i=1,2,3,\ldots,242$, we select the two most correlated animals
in the $i$'th dish, and the first scientist picks---equiprobably---one of
these two animals, and sets ${\bf X}^{(i)}\in \R^{16}$ to be this animal's
associated vector, and the other scientist is left with the other animal,
and sets  ${\bf Y}^{(i)}\in \R^{16}$ to be that animal's associated vector.
The first scientist's observations are stored in the matrix
$X^{(242)}=[ {\bf X}^{(1)}|{\bf X}^{(2)}|\cdots |{\bf X}^{(242)} ] \in \R^{16 \times 242}$
and the second scientist's observations are stored in the matrix
$Y^{(242)}=[ {\bf Y}^{(1)}|{\bf Y}^{(2)}|\cdots |{\bf Y}^{(242)} ] \in \R^{16 \times 242}$.
(When we replicate our experiment, the identities of the
two most correlated animals in each dish don't change from one
experiment replication to the next, but which of the two animals
is assigned to the first scientist are independent Bernoulli$(\frac{1}{2})$ trials.)

For each embedding dimension $k=1,2,\ldots, 16$, we use PCA to generate
${\mathcal A}^{(242)}$ and ${\mathcal B}^{(242)}$, and then we compute
$d^2({\mathcal A}^{(242)},{\mathcal B}^{(242)})$ and
$\epsilon^2({\mathcal X}^{(242)},{\mathcal Y}^{(242)})$ in the
manner described in Section~\ref{setting}. Performing $10000$ Monte-Carlo replications
of this experiment, we plot in Figure \ref{fignice} the values of
$\epsilon^2({\mathcal X}^{(242)},{\mathcal Y}^{(242)})$
against the values of $d^2({\mathcal A}^{(242)},{\mathcal B}^{(242)})$
for each of these $10000$ replicates, and for each of the embedding dimensions
$k=1,2, \ldots, 16$. The colors of the plotted points are
red, green, blue, black, cyan, magenta,
red, green, blue, black, cyan, magenta,
red, green, blue, black according as the embedding
dimension is $k=1,2,3,\ldots, 16$.
Note that for each embedding dimension $k=1,2,\ldots, 15$,
there is a positive linear correlation in the
the plotted points of Figure~\ref{fignice}.
In particular, note the substantial increase in the standard
deviation of $d^2 ({\mathcal A}^{(242)},{\mathcal B}^{(242)})$
as embedding dimension changed from $k=1$ to $k=2$ to
$k=3$, and again from $k=6$ to $k=7$. Although the (normalized)
values of $\epsilon^2({\mathcal X}^{(242)},{\mathcal Y}^{(242)})$
seem to be anyway increasing as $k$ increases, it also seems that
increases in $\epsilon^2({\mathcal X}^{(242)},{\mathcal Y}^{(242)})$
are also explained by the increased values of
$d^2 ({\mathcal A}^{(242)},{\mathcal B}^{(242)})$, as these
increased values of $d^2 ({\mathcal A}^{(242)},{\mathcal B}^{(242)})$
occur. This is the incommensurability phenomenon. Although it is not
as dramatic as with the simulated data, it is present
in this real-data setting.

For an instantiation of one of the scientist's data,
the sample covariance matrix had eigenvalues
$.06284$, $.01896$, $.00988$, $.00748$, $.00618$, $.00473$,
$.00328$, $.00312$, $.00291$, $.00254$, $.00244$, $.00228$,
$.00185$, $.00162$, $.00140$, $.00128$. Note that there was a
precipitous narrowing of eigengap between the $7$th eigenvalue
and the $8$th eigenvalue; this corresponds to the sudden change
in behavior in Figure \ref{fignice} between embedding dimension
$k=6$ and embedding dimension $k=7$.

\section{Summary and discussion}

When principal components analysis (PCA) is used for the dimension
reduction of two random data sets that are highly-correlated
with each
other, there is a natural hope that that the projected (and normalized)
data sets will be commensurate, in the sense that a Procrustes transformation
of one to the other will render it close in distance (according to the
strength of correlation in the original data). However, sometimes this
Procrustean fitting-error is higher than what might be expected, which
is the ``incommensurability phenomenon." This may occur when
the projections are done separately for the two data sets and there is
an insufficient gap between covariance eigenvalues as the more-principal
principal components are taken and less-principal principal components
are discarded, which can lead to nontrivial variance in the
resulting PCA projectors. (Indeed, the Cautionary Tale of Two Scientists 
from Section \ref{caut}, with spherical covariance structure, creates 
a perfect storm.)

Our main result is Theorem \ref{realmain}, which succinctly
quantifies the asymptotic effects of (an adaption of) the Hausdorff distance
between the PCA projections, in terms of the strength of the
correlation between the original data sets, on the Procrustean
fitting-error of the projected data.
We then illustrate that highly-correlated data,
even with a mild gap in covariance eigenvalues, can appreciably
exhibit the incommensurability phenomenon; indeed, what we observe
from the simulations is very closely aligned with the asymptotic
relationship that we proved.

Awareness of these results is important when decisions are made
regarding dimension reduction for separate data sets assumed to
represent similar phenomena. For example, in distributed settings it
may be assumed that highly correlated large data sets can be merged
{\it after} dimension reduction, thereby allowing for more computationally
efficient data transfer. However, our results indicate that this
approach can be disastrous, even when the assumption that the separate
data sets are highly correlated is valid.
\\ \\

\noindent
{\bf Acknowledgements:}
 The work of all authors was partially supported
by National Security Science and Engineering Faculty Fellowship (NSSEFF),
Johns Hopkins University Human Language Technology
Center of Excellence (JHU HLT COE), and the XDATA program of the
Defense Advanced Research Projects Agency (DARPA) administered
through Air Force Research Laboratory contract FA8750-12-2-0303.
The authors are grateful to the anonymous referees for particularly
useful comments that greatly strengthened this paper.

\end{document}